\documentclass[]{IEEEtran}
\usepackage[dvips,]{graphicx}
\usepackage{amssymb,amsmath,array}
\usepackage{dsfont}
\usepackage{psfrag} % for postscript graphics files!!!!!!ÓÃÀ´Ìæ»»Í¼Æ¬ÄÚÈÝ
\usepackage{amsmath,amsthm} % assumes amsmath package installed
\usepackage[x11names]{xcolor}
\usepackage{overpic}
\usepackage{multirow}
\usepackage{rotating}
\usepackage{overpic}
\usepackage{hyperref}
\usepackage{makecell}%%%%%%%%%%%%%%%% set linewidth in the table
\usepackage{enumerate}
\usepackage{algorithm}

\usepackage{algorithmic}
\usepackage{array} % ?????§á?
\usepackage{subfigure}
\usepackage{color}
\usepackage[utf8]{inputenc} % Pour le français

\newcommand{\cb}[1]{\ifmmode {\boldsymbol{#1}}\else ${\boldsymbol{#1}}$\fi}
\newcommand{\cp}[1]{\ifmmode {\mathcal{#1}}\else ${\mathcal{#1}}$\fi}
\onecolumn
\def\E{\text{I}\!\text{E}}

\newcommand{\bA}{\cb{A}}
\newcommand{\bB}{\cb{B}}

\newcommand{\bx}{\cb{x}}
\newcommand{\bz}{\cb{z}}
\newcommand{\bu}{\cb{u}}
\newcommand{\by}{\cb{y}}
\newcommand{\bX}{\cb{X}}
\newcommand{\bY}{\cb{Y}}
\newcommand{\bM}{\cb{M}}
\newcommand{\bm}{\cb{m}}
\newcommand{\bn}{\cb{n}}

\newcommand{\bc}{\cb{c}}

\begin{document}
%
% paper title
% can use linebreaks \\ within to get better formatting as desired
% Do not put math or special symbols in the title.
\title{Correntropy Maximization via ADMM \\-- Application to Robust Hyperspectral Unmixing --}%Correntropy Maximization}% with ADMM for %alternating direction method of multipliers
%
% author names and IEEE memberships
% note positions of commas and nonbreaking spaces ( ~ ) LaTeX will not break
% a structure at a ~ so this keeps an author's name from being broken across
% two lines.
% use \thanks{} to gain access to the first footnote area
% a separate \thanks must be used for each paragraph as LaTeX2e's \thanks
% was not built to handle multiple paragraphs
%

\author{Fei~Zhu, Abderrahim~Halimi, %~\IEEEmembership{Member,~IEEE},
        Paul~Honeine, %~\IEEEmembership{Member,~IEEE}, 
        Badong~Chen, %~\IEEEmembership{Senior Member,~IEEE},~
        Nanning~Zheng%,%~\IEEEmembership{Fellow,~IEEE}
\thanks{F. Zhu is with the Institut Charles Delaunay (CNRS), Universit\'{e} de Technologie de Troyes, France. (fei.zhu@utt.fr)}
\thanks{A. Halimi is with the School of Engineering and Physical Sciences, Heriot-Watt University, U.K. (a.halimi@hw.ac.uk)}
\thanks{P. Honeine is with the LITIS lab, Universit\'{e} de Rouen, France. (paul.honeine@univ-rouen.fr)}
\thanks{Badong Chen and Nanning Zheng are with the Institute of Artificial Intelligence and Robotics, Xi'an Jiaotong University, Xi'an, China. (chenbd;~nnzheng@mail.xjtu.edu.cn)
}
}

\maketitle

% As a general rule, do not put math, special symbols or citations
% in the abstract or keywords.
\begin{abstract}
In hyperspectral images, some spectral bands suffer from low signal-to-noise ratio due to noisy acquisition and atmospheric effects, thus requiring robust techniques for the unmixing problem. This paper presents a robust supervised spectral unmixing approach for hyperspectral images. The robustness is achieved by writing the unmixing problem as the maximization of the correntropy criterion subject to the most commonly used constraints. Two unmixing problems are derived: the first problem considers the fully-constrained unmixing, with both the non-negativity and sum-to-one constraints, while the second one deals with the non-negativity and the sparsity-promoting of the abundances. The corresponding optimization problems are solved efficiently using an alternating direction method of multipliers (ADMM) approach. Experiments on synthetic and real hyperspectral images validate the performance of the proposed algorithms for different scenarios, demonstrating that the correntropy-based unmixing is robust to outlier bands.% It is demonstrated that the correntropy based approach is robust to the outlier bands (such as the water absorption bands).

\end{abstract}

% Note that keywords are not normally used for peerreview papers.

% For peer review papers, you can put extra information on the cover
% page as needed:
% \ifCLASSOPTIONpeerreview
% \begin{center} \bfseries EDICS Category: 3-BBND \end{center}
% \fi
%
% For peerreview papers, this IEEEtran command inserts a page break and
% creates the second title. It will be ignored for other modes.
\IEEEpeerreviewmaketitle

%\bigskip

\newpage

\section{Introduction}
% The very first letter is a 2 line initial drop letter followed
% by the rest of the first word in caps.
%
% form to use if the first word consists of a single letter:
% \IEEEPARstart{A}{demo} file is ....
%
% form to use if you need the single drop letter followed by
% normal text (unknown if ever used by IEEE):
% \IEEEPARstart{A}{}demo file is ....
%
% Some journals put the first two words in caps:
% \IEEEPARstart{T}{his demo} file is ....
%
% Here we have the typical use of a "T" for an initial drop letter
% and "HIS" in caps to complete the first word.
\IEEEPARstart{S}pectral unmixing is an essential issue in many disciplines, including signal and image processing, with a wide range of applications, such as classification, segmentation, material identification and target detection. Typically, a hyperspectral image corresponds to a scene taken at many continuous and narrow bands across a certain wavelength range; namely, each pixel is a spectrum. Assuming that each spectrum is a mixture of several pure materials, the unmixing problem consists in two tasks: ($\text{\romannumeral1}$) identifying these pure materials (the so-called \emph{endmembers}); ($\text{\romannumeral2}$) estimating their proportions (the so-called \emph{abundances}) at each pixel~\cite{Keshava02}. In practice, these two steps can be performed either sequentially or simultaneously~\cite{12.tgrs.barycenters}. Well-known endmember extraction algorithms include the pure-pixel-based ones, {\em e.g.}, the vertex component analysis (VCA)~\cite{VCA} and the N-FINDR~\cite{N-FINDR}, as well as the minimum-volume-based ones, {\em e.g.}, the minimum simplex analysis~\cite{MinimumSimplex} and the minimum volume constrained nonnegative matrix factorization~\cite{Miao07}. While the endmember extraction is relatively easy from geometry, the abundance estimation remains an open problem.
%Indeed, besides estimating the abundances from least-squares methods or from geometry~\cite{12.tgrs.barycenters}, one can tackle recently-raised issues such as nonlinearity~\cite{13.tsp.unmix,14.tgrs.nonlinear}.
Indeed, the abundances can be estimated using least-squares methods, geometric approaches~\cite{12.tgrs.barycenters}, or by tackling recently-raised issues such as nonlinearity~\cite{13.tsp.unmix,14.tgrs.nonlinear}.
In this paper, we consider the abundance estimation problem.%in unmixing hyperspectral images.

%By supposing that the endmembers are available using any endmember extraction algorithm, the problem of estimating the abundances is often referred as the ``supervised mixing problem''.

%Referred by ``supervised'', this paper is restricted to the abundance estimation algorithm,

%Relying on different physical assumptions, two categories of mixture models exist, namely the linear and nonlinear ones. %Compared to its nonlinear counterparts,
The linear mixture model (LMM) is the most investigated over the past decades~\cite{Miao07, Heinz, Huck}.
%Besides the simplicity, it is also because that the light scattering mechanisms approximated by LMM is often acceptable \cite{bioucas2012hyperspectral}.
Its underlying premise is that each pixel/spectrum is a linear combination of the endmembers. To be physically interpretable, two constraints are  often enforced in the estimation problem: the abundance non-negativity constraint (ANC) and the abundance sum-to-one constraint (ASC) for each pixel \cite{bioucas2012hyperspectral}. Considering both constraints, the fully-constrained least-squares method (FCLS) was presented in~\cite{Heinz}. A more recently proposed unmixing algorithm is the so-called SUnSAL, for {\em Sparse Unmixing by variable Splitting and Augmented Lagrangian} \cite{bioucas2010alternating}. It addresses the same optimization problem by taking advantage of the alternating direction method of multipliers (ADMM)~\cite{boyd2011distributed}. A constrained-version of SUnSAL was also proposed to solve the constrained sparse regression problem, where the sum-to-one constraint (ASC) is relaxed and the $\ell_1$-norm regularizer is added.

All these unmixing algorithms hugely suffer from noisy data and outliers within bands. Indeed, % Unsupervised Correntropy unmixing
in real hyperspectral images for remote sensing, a considerable proportion (about 20\%) of the spectral bands are noisy with low SNR, due to the atmospheric effect such as water absorption~\cite{Zelinski06}. These bands need to be removed prior to applying any existing unmixing method; otherwise, the unmixing quality drastically decreases. Such sensitivity to outliers is due to the investigated $\ell_2$-norm as a cost function in the FCLS and SUnSAL algorithms, as well as all unmixing algorithms that explore least-squares solutions. It is worth noting that nonlinear unmixing algorithms also suffer from this drawback, including the kernel-based fully-constrained least-squares (KFCLS) \cite{broadwater2007kernel}, nonlinear fluctuation methods \cite{13.tsp.unmix} and post-nonlinear methods \cite{13.whispers.postnonlinear}.

%Most existing unmixing methods %, regardless linear~\cite{Iordache11, Heinz} or nonlinear~\cite{HaAlDoTo2011, 13.tsp.unmix},
%are sensitive to the outliers and need to be performed on the reduced data after removing these noisy bands; Otherwise,

Information theoretic learning provides an elegant alternative to the conventional minimization of the $\ell_2$-norm in least-squares problems, by considering the maximization of the so-called correntropy~\cite{liu2007correntropy, principe2010information}. Due to its stability and robustness to noise and outliers, the correntropy maximization is based on theoretical foundations and has been successfully applied to a wide class of applications, including cancer clustering~\cite{wang2013non}, face recognition~\cite{he2011maximum}, and recently hyperspectral unmixing \cite{WangCorrentropy}, to name a few. In these works, the resulting problem is optimized by the half-quadratic technique~\cite{nikolova2005analysis}, either in a supervised manner~\cite{he2011maximum} or as an unsupervised nonnegative matrix factorization~\cite{wang2013non, WangCorrentropy}. %{\magenta Unfortunately, the half-quadratic technique is not suitable for nonconvex problems, such as with the correntropy maximization. ???? so as ADMM...}%The non-negativity constraint is imposed and the sparsity is augmented by the $\ell_1$-norm regularizer.

In this paper, we consider the hyperspectral unmixing problem by defining an appropriate correntropy-based criterion, thus taking advantage of its robustness to large outliers, as opposed to the conventional $\ell_2$-norm criteria. By including constraints commonly used for physical interpretation, we propose to solve the resulting constrained optimization problems with alternating direction method of multipliers (ADMM) algorithms. Indeed, the ADMM approach splits a hard problem into a sequence of small and handful ones \cite{boyd2011distributed}. Its relevance to solve nonconvex problems was studied in \cite[Section 9]{boyd2011distributed}. We show that ADMM provides a relevant framework for incorporating different constraints raised in the unmixing problem. We present the so-called CUSAL (for {\em Correntropy-based Unmixing by variable Splitting and Augmented Lagrangian}), and study in particularly two algorithms: CUSAL-FC to solve the fully-constrained (ANC and ASC) correntropy-based unmixing problem, and the CUSAL-SP to solve the sparsity-promoting correntropy-based unmixing problem.

The rest of the paper is organized as follows. We first provide a succinct survey on the classical unmixing problems in Section~\ref{sec: LMM}. In Section~\ref{sec:Correntropy}, we propose the correntropy-based unmixing problems subject to the aforementioned constraints, and study the robustness. The resulting optimization problems are solved by the ADMM algorithms described in Section~\ref{sec:Algorithm}. Experiments on synthetic and real hyperspectral images are presented in Sections~\ref{sec: Synthetic} and~\ref{sec: Real}, respectively.  Finally, Section~\ref{sec: Conclusion} provides some conclusions and future works.

\section {Classical Unmixing Problems}\label{sec: LMM}

%{\blue La notation est \`a d\'efinir. Je pense qu'on peut faire mieux, parce que $\by^l$ n'est pas g\'enial. On peut faire $y_{lt}$, $\by_{*t}$ et $\by_{l*}$, pour d\'esigner un \'el\'ement, une colonne $t$ et une ligne $l$ de $\bY$, respectivement. On d\'efinit AUSSI $\by_{t}=\by_{*t}$. Comme ca, il y le moins de modifications \`a faire, puisque \eqref{eq:LMM} reste correcte.}
The linear mixture model (LMM) assumes that %each arriving photon interacts with just one pure substance, thus
each spectrum can be expressed as a linear combination of a set of pure material spectra, termed endmembers~\cite{Keshava02}.
Consider a hyperspectral image and let $\bY \in \mathds{R}^{L\times T}$ denote the matrix of the $T$ pixels/spectra of $L$ spectral bands. Let $\by_{*t}$ be its $t$-th column and $\by_{l*}$ its $l$-th row, representing the $l$-th band of all pixels. For notation simplicity, we denote $\by_t=\by_{*t}$, for $t=1,\ldots, T$.
%$\by_t \in \mathds{R}^{L}$, for $t=1,\ldots, T$, and let $\bY=[ \by_{*1} ~ \cdots ~ \by_{*T} ]= [ \by_{1*} ~ \cdots ~ \by_{L*} ]^{\top}\in \mathds{R}^{L\times T}$
The LMM can be written as
\begin{equation}\label{eq:LMM}
  \by_{t}=\sum_{r=1}^R x_{rt} \, \bm_r +\bn_t=\bM\bx_t+\bn_t,
\end{equation}
where $\bM =[ \bm_1 ~ \cdots ~ \bm_R ]\in \mathds{R}^{L\times R}$ is the matrix composed by the $R$ endmembers with $\bm_r=[m_{1r} ~ \cdots ~ m_{Lr}]^\top$, $\bx_t=[x_{1t} ~ \cdots ~ x_{Rt}]^\top$ is the abundance vector associated with the $t$-th pixel, and $\bn_t \in \mathds{R}^{L}$ is the additive noise. In matrix form for all pixels, we have
%\begin{equation*}%\label{eq:LMM_matrix}
$  \bY=\bM\bX+\cb{N}$, 
%\end{equation*}
where $\bX=[ \bx_1 ~ \cdots ~ \bx_T ] \in\mathds{R}^{R\times T}$ and $\cb{N}$ is the noise matrix.

In the following, the endmembers are assumed known, either from ground-truth information or by using any endmember extraction technique. The spectral unmixing problem consists in estimating the abundances for each pixel, often by solving the least-squares optimization problem
\begin{equation}\label{eq:ls_prob}
	\min_{\bx_t} \|\by_t - \bM\bx_t \|_2^2,
\end{equation}
for each $t =1, \ldots, T$, where $\|\cdot\|_2$ denotes the conventional $\ell_2$-norm. The solution to this conventional least-squares problem is given by the pseudo-inverse of the (tall) endmember matrix, with $\bx_t = (\bM^\top \bM)^{-1} \bM^\top \by_t$. The least-squares optimization problems \eqref{eq:ls_prob}, for all $t=1, \ldots, T$, are often written in a single optimization problem using the following matrix formulation
\begin{equation}\label{eq:ls_prob_matrix}
	\min_{\bX} \|\bY - \bM\bX \|_F^2,
\end{equation}
where $\| \cdot \|_F^2$ denotes the Frobenius norm. Its solution is
\begin{equation}\label{eq:ls_solution}
\bX_\text{LS} = (\bM^\top \bM)^{-1} \bM^\top \bY.
\end{equation}
Finally, this optimization problem can be also tackled by considering all the image pixels at each spectral band, which yields the following least-squares optimization problem
\begin{equation*}%\label{eq:LS}
 \min_{\bX} \sum_{l=1}^L \|\by_{l*} - (\bM\bX)_{l*} \|_2^2,
\end{equation*}
where $(\cdot)_{l*}$ denotes the $l$-th row of its argument. While all these problem formulations have a closed-form solution, they suffer from two major drawbacks. The first one is that several constraints need to be imposed in order to have a physical meaning of the results. The second drawback is its sensitivity to noise and outliers, due to the use of the $\ell_2$-norm as a fitness measure. These two drawbacks are detailed in the following.%To overcome this difficulty, we provide the correntropy measure for robust solutions, as given in Section~\ref{sec:Correntropy}, while algorithms to maximizing the constrained correntropy are derived in Section~\ref{sec:Algorithm}.

%\subsection{Constraints}

To be physically interpretable% as emphasized in~\cite{Keshava02}
, the abundances should be nonnegative (ANC) and satisfy the sum-to-one constraint (ASC). Considering both constraints, the fully-constrained least-squares problem is formulated as, for each $t =1, \ldots, T$,
\begin{align*}
\begin{aligned}
\displaystyle \min_{\bx_t}  \|\by_t - \bM\bx_t \|_2^2, %\\
\text{ subject to } {\bx_t} \succeq 0 \text{ and } {\cb1^{\top}\bx_t= 1},%, \text{for } t =1, \ldots, T,
\end{aligned}
\end{align*}
where $\cb1\in\mathds{R}^{R\times 1}$ denotes the column vector of ones and $\succeq0$ is the non-negativity applied element-wise; In matrix form:
\begin{align*}
\begin{aligned}
\displaystyle \min_{\bX} \|\bY - \bM\bX \|_F^2, %\\
\text{ subject to } & {\bX} \succeq 0 \\
\text{ and } & {\cb1^{\top}} \bx_t= 1, \text{ for } t =1, \ldots, T.
\end{aligned}
\end{align*}
Since there is no closed-form solution when dealing with the non-negativity constraint, several iterative techniques have been proposed, such as the {\em active set} scheme with the Lawson and Hanson's algorithm \cite{LS87}, the multiplicative iterative strategies \cite{Lanteri2001}, and the fully-constrained least-squares (FCLS) technique \cite{Heinz}. More recently, the alternating direction method of multipliers (ADMM) was applied with success for hyperspectral unmixing problem, with the SUnSAL algorithm~\cite{bioucas2010alternating}.

Recent work in hyperspectral unmixing  have advocated the sparsity of the abundance vectors~\cite{bioucas2010alternating, Iordache11, Iordache2012}. In this case, each spectrum is fitted by a sparse linear mixture of endmembers, namely only the abundances with respect to a small number of endmembers are nonzero. To this end, the sparsity-promoting regularization with the $\ell_1$-norm is included in the cost function, yielding the following constrained sparse regression problem~\cite{bioucas2010alternating}, for each $t =1, \ldots, T$,
\begin{align*}
\begin{aligned}
\displaystyle \min_{\bx_t} \|\by_t - \bM\bx_t \|_2^2 + \lambda \|\bx_t\|_1, %\\
\text{ subject to } {\bx_t} \succeq 0,%, \text{ for }~ t =1, \ldots, T.
\end{aligned}
\end{align*}
where the parameter $\lambda$ balances the fitness of the least-squares solution and the sparsity level. It is worth noting that the ASC is relaxed when the $\ell_1$-norm is included. This problem is often considered by using the following matrix formulation
\begin{align*}
\begin{aligned}
\displaystyle \min_{\bX} \|\bY - \bM\bX \|_F^2 + \lambda \sum_{t=1}^T \|\bx_t\|_1,% \\
\text{ subject to } {\bX} \succeq 0.%,%, \text{ for }~ t =1, \ldots, T.
\end{aligned}
\end{align*}
%or by substituting the sparsity-promoting term with the $\ell_1$-norm of the matrix $\bX$.

\subsection*{Sensitivity to outliers}

All the aforementioned algorithms rely on solving a (constrained) least-squares optimization problem, thus inheriting the drawbacks of using the $\ell_2$-norm as the fitness measure. A major drawback is its sensitivity to outliers, where outliers are some spectral bands that largely deviate from the rest of the bands. Indeed, considering all the image pixels, the least-squares optimization problems take the form
\begin{equation}\label{eq:LS}
 \min_{\bX} \sum_{l=1}^L \|\by_{l*} - (\bM\bX)_{l*} \|_2^2,
\end{equation}
subject to any of the aforementioned constraints. From this formulation, it is easy to see how the squared $\ell_2$-norm gives more weight to large residuals, namely to outliers in which predicted values $(\bM\bX)_{l*}$ are far from actual observations $\by_{l*}$. Moreover, it is common for hyperspectral images to present up to 20\% of unusable spectral bands due to low signal-to-noise ratio essentially from atmospheric effects, such as water absorption. In the following section, we overcome this difficulty by considering the correntropy maximization principle from the information theoretic learning, which yields an optimization problem that is robust to outliers.

% each spectral band contributes uniformly to the residual error. Moreover,
%This may be helpful in studies where outliers do not need to be given greater weight than other observations

\section{Correntropy-based Unmixing Problems}\label{sec:Correntropy}

In this section, we examine the correntropy and write the %hyperspectral 
unmixing problems as correntropy maximization ones. Algorithms for solving these problems are derived in Section~\ref{sec:Algorithm}.

\subsection {Correntropy}
 %Il manque une explication sur la correntropie, et les origines de la correntropie. C'est mieux d'expliquer, au moins un peu.
The correntropy, studied in~\cite{liu2007correntropy, principe2010information}, is a nonlinear local similarity measure. % between two arbitrary random variables.
For two random variables, $\mathcal{Y}$ and its estimation $\widehat{\mathcal{Y}}$ using some model/algorithm, it is defined by
\begin{equation}\label{eq:Correntropy}
\E[\kappa(\mathcal{Y},\widehat{\mathcal{Y}})],
\end{equation}
where $\E[\cdot]$ is the expectation operator, and $\kappa(\cdot,\cdot)$ is a shift-invariant kernel satisfying the Mercer theorem \cite{Vap95}.
In practice, while the joint distribution function of $\mathcal{Y}$ and $\widehat{\mathcal{Y}}$ is unavailable, the sample estimator of correntropy is adopted instead. Employing a finite number of data $\{(\by_{l*}, \widehat{\by}_{l*})\}_{l=1}^L$, it is estimated by
\begin{equation}\label{eq:CorrentropySampleEstimator}
\frac{1}{L} \sum_{l=1}^L \kappa(\by_{l*},\widehat{\by}_{l*}),
\end{equation}
up to a normalization factor. The Gaussian kernel is the most commonly-used kernel for correntropy~\cite{liu2007correntropy, he2011maximum, chen2012maximum}. %It is defined by $\kappa(u,v) = \exp (\frac{-1}{{2} \sigma^2} \|u- v\|^2)$, where $\sigma$ denotes the kernel bandwidth.
%{\magenta The Gaussian kernel can be interpreted as a similarity measure between $u$ and $v$, that equals to 1 if $u=v$, and tends to 0 as two arguments are very distant.}
This leads to the following expression for the correntropy
\begin{equation}\label{eq:Correntropy_Gaussian}
  \frac{1}{L} \sum_{l=1}^{L} \exp\left(\tfrac{-1}{2\sigma^2} \| \by_{l*} - \widehat{\by}_{l*}\|_2^2 \right),
\end{equation}
where $\sigma$ denotes the bandwidth of the Gaussian kernel.
%It is clearly shown that reducing the distance between $u_l$ and $v_l$ is equivalent to the maximzation of the correntropy criterion.

The maximization of the %similarity measured by the 
correntropy, given by
\begin{equation*}%\label{eq:CorrentropySampleEstimator}
\max_{\widehat{\by}_{1*}, \ldots, \widehat{\by}_{L*}} \frac{1}{L}\sum_{l=1}^L \kappa(\by_{l*},\widehat{\by}_{l*}),
\end{equation*}
is termed the maximum correntropy criterion \cite{liu2007correntropy}.
It is noteworthy that well-known second-order statistics, such as the mean square error (MSE) depends heavily on the Gaussian and linear assumptions~\cite{liu2007correntropy}.
However, in presence of non-Gaussian noise and in particular large outliers, {\em i.e.}, observations greatly deviated from the data bulk, the effectiveness of the MSE-based algorithms will significantly deteriorate~\cite{Wu15}. By contrast, the maximization of the correntropy criterion is appropriate for non-Gaussian signal processing, and is robust in particular against large outliers, as shown next.%, since the correntropy is a localized similarity~\cite{liu2007correntropy}.

%{\red \bigskip
%Il faut expliquer pourquoi on maximize la correntropie.
%}
\subsection{The underlying robustness of the correntropy criterion}

\begin{figure}
\centering
\graphicspath{{Graphics/}}
\centering
\psfragscanon
\psfrag{-6}{}%\!\!\tiny$_{-6}$}
\psfrag{-5}{\!\!\tiny$_{-5}$}
\psfrag{-4}{\!\!\tiny$_{-4}$}
\psfrag{-3}{\!\!\tiny$_{-3}$}
\psfrag{-2}{\!\!\tiny$_{-2}$}
\psfrag{-1}{\!\!\tiny$_{-1}$}
\psfrag{6}{\!\tiny$_{6}$}
\psfrag{0}[c][c]{\!\tiny$_{0}$}
\psfrag{5}{\!\tiny$_{5}$}
\psfrag{4}{\!\tiny$_{4}$}
\psfrag{3}{\!\tiny$_{3}$}
\psfrag{2}{\!\tiny$_{2}$}
\psfrag{1}{\!\tiny$_{1}$}
\psfrag{0.5}{\!\!\!\!\tiny$_{0.5}$}
\psfrag{1.5}{\!\!\!\!\tiny$_{1.5}$}
%\psfrag{2.5}{}
%\psfrag{3.5}{}
%\psfrag{4.5}{}
%\psfrag{5.5}{}
%\psfrag{outliers}{\!\!\!\!\:\tiny outliers}
\psfrag{residual error}[c][c]{\scriptsize residual error}% $\epsilon_l$}
\psfrag{s}{}
\psfrag{= 0.5 }{\small correntropy ($\sigma=0.5$)}
\psfrag{= 2 }{\small correntropy ($\sigma=2$)}
\psfrag{=  5 }{\small correntropy ($\sigma=5$)}
\psfrag{least square}{\small second-order function}

\includegraphics[trim = 40mm 0mm 30mm 0mm, clip,width=0.9\textwidth,]{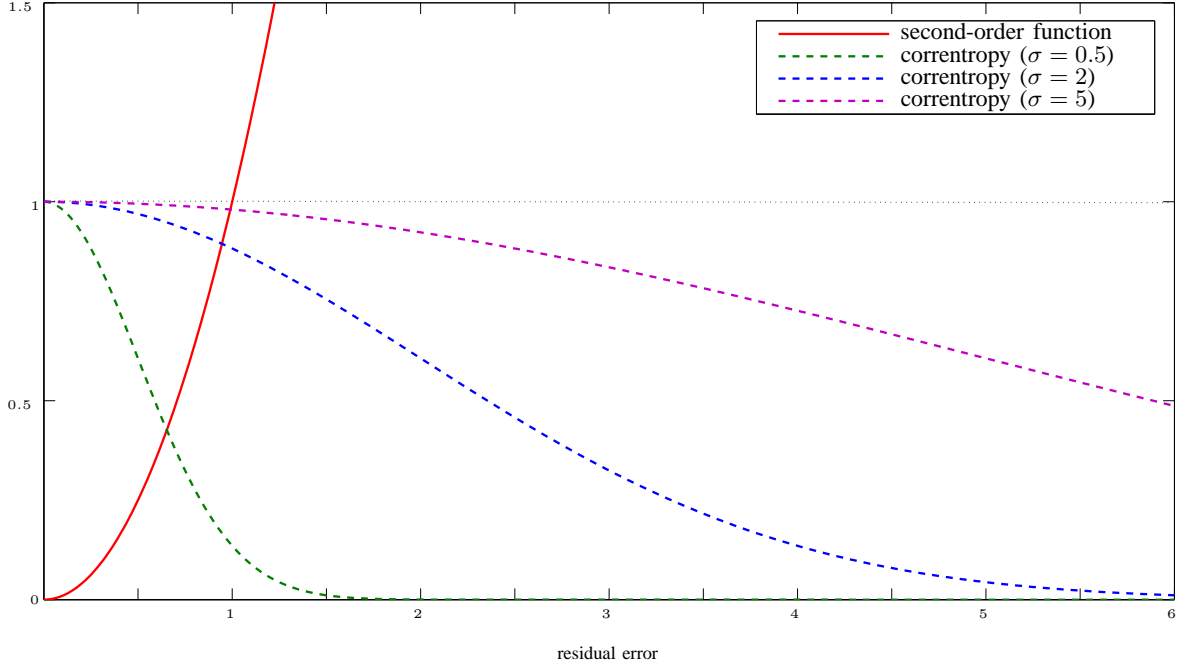}
\caption{Illustration of second-order objective function (solid line) and the correntropy objective function (dashed lines) with different values of the kernel bandwidth.}
\label{Fig. CorrenLS}
\end{figure}

In this section, we study the sensitivity to outliers of the correntropy maximization principle, by showing the robustness of the underlying mechanism. To this end, we examine the behavior of the correntropy in terms of the residual error defined by
%\begin{equation*}
%	\epsilon(l) = \|u_l - v_l\|.
%\end{equation*}
%Therefore, the correntropy \eqref{eq:Correntropy_Gaussian} becomes
%\begin{equation*}
%	\frac{1}{L} \sum_{l=1}^{L} \exp\left(\tfrac{-1}{2\sigma^2} \epsilon(l)^2 \right)
%\end{equation*}
%The derivative of this objective function with respect to any residual error $\epsilon(l)$ is
%\begin{equation*}
%	-\frac{1}{L \sigma^2} \exp\left(\tfrac{-1}{{2} \sigma^2} \epsilon(l)^2 \right) \, \epsilon(l).
%\end{equation*}
%{\red \figurename~\ref{fig:correntropy}...}
%\begin{figure}[t]
%\centering
%\psfragscanon
%\psfrag{correntropy .1}{\tiny correntropy $\sigma=0.1$}
%\psfrag{correntropy .5}{\tiny correntropy $\sigma=.5$}
%\psfrag{correntropy 1}{\tiny correntropy $\sigma=1$}
%\psfrag{least-squares               }{\tiny least-squares}
%\psfrag{x}[c][c]{\footnotesize residual error}
%\psfrag{y}[c][c]{\footnotesize derivatives}
%
%\includegraphics[width=0.5\textwidth]{correntropy.eps}
%\caption{\red ???}
%\label{fig:correntropy}
%\end{figure}
%This section studies the sensitivity to outliers of the MCC, by showing the robustness of the underlying mechanism.
%We examine the behavior of the correntropy in terms of the residual error defined by
%\begin{equation*}
$\epsilon_l \!=\! \|\by_{l*} \!-\! \widehat{\by}_{l*}\|_2$. 
%\end{equation*}
Thus, the correntropy \eqref{eq:Correntropy_Gaussian} becomes
\begin{equation*}
\frac{1}{L} \sum_{l=1}^{L} \exp\left(\tfrac{-1}{2\sigma^2} \epsilon_l^2 \right).
\end{equation*}
Compared with second-order statistics, {\em e.g.} MSE, the correntropy is more robust with respect to the outliers, as shown in \figurename~\ref{Fig. CorrenLS} illustrating the second-order and the correntropy objective functions in terms of the residual error.
As the residual error increases, the second-order function keeps increasing dramatically. On the contrary, the correntropy is only sensitive within a region of small residual errors, this region being controlled by the kernel bandwidth. For large magnitudes of residual error% out of the central range
, the correntropy falls to zero%grows extremely gently and tends to a constant
. Consequently, the correntropy criterion is robust to large outliers.%, especially for small kernel bandwidths.

\subsection{Correntropy-based unmixing problems}

%Assuming a known endmember matrix $\bM$,
The correntropy-based unmixing problem consists in estimating the unknown abundance matrix $\bX$, by minimizing the objective function $\mathcal{C}$ (the negative of correntropy), given by
\begin{equation}\label{eq:ObjectiveCorrentropy}
  \mathcal{C}(\bX) = -\sum_{l=1}^{L} \exp\left(\tfrac{-1}{2\sigma^2} \|\by_{l*} - (\bM\bX)_{l*}\|_2^2 \right) ,
\end{equation}
where the Gaussian kernel was considered% and the nonnegative constraint is enforced on all the entries of $\bX$
, or equivalently
\begin{equation}\label{eq:ObjectiveCorrentropy2}
  \mathcal{C}(\bX) = -\sum_{l=1}^{L} \exp \left(\tfrac{-1}{2\sigma^2} \sum_{t=1}^T \Big(y_{lt} - \sum_{r=1}^R x_{rt} \, m_lr \Big)^2 \right).
\end{equation}
%In hypersepctral unmixing problem, the sum-to-one constraint is often required, that is, for each pixel, the abundances contributed by all the endmembers should be added to one.

Considering both the ANC and ASC constraints, the fully-constrained correntropy unmixing problem becomes
\begin{align}\label{eq:SumToOne}
\begin{aligned}
\displaystyle \min_{\bX}~  \mathcal{C}(\bX), %\\
\text{ subject to } & {\bX} \succeq 0 \\
\text{ and } & {\cb1^\top\bx_t= 1}, \text{ for } t =1, \ldots, T.
\end{aligned}
\end{align}
%This optimization problem is termed CUSAL-FC for correntropy-based unmixing with full-constraints.
%where $\cb1\in\mathds{R}^{R\times 1} $ is the vector of one.
%\subsection{Sparsity}
For the sake of promoting sparse representations, the objective function~\eqref{eq:ObjectiveCorrentropy}-\eqref{eq:ObjectiveCorrentropy2} can be augmented by the $\ell_1$-norm penalty on the abundance matrix $\bX$, leading to the following problem
\begin{align}\label{eq:Sparsity}
\begin{aligned}
\displaystyle &\min_{\bX} \mathcal{C}(\bX)+ \lambda \sum_{t=1}^{T}\|\bx_t\|_1, %\\
\text{ subject to } {\bX} \succeq 0.
\end{aligned}
\end{align}
%where the parameter $\lambda$ balances the fitness of the solution to the data fidelity term and the sparsity level.

%
%{\red
%It is worth noting that the mixture model with correntropy is related to the kernel, nonlinear method. Indeed, the correntropy can be viewed as a generalized correlation function defined in some nonlinear, high-dimensional feature space~\cite{liu2007correntropy}. %From the perspective of kernel method,
%Denote
%$\widehat{\by}_{l*}=(\bM\bX)_{l*}$ for notation simplicity, the problem of minimizing~\eqref{eq:ObjectiveCorrentropy} is rewritten as
%\begin{equation*}
%	\min_{\bX} -\sum_{l=1}^L \kappa(\by_{l*},\widehat{\by}_{l*}).
%\end{equation*}
%Since $\kappa$ is a Mercer kernel, it corresponds to a nonlinear function $\Phi(\cdot)$ that maps all the $\by_{l*}$ and $\widehat{\by}_{l*}$ from the input space to a high-dimensional feature space $\cp{H}$. Thus, the optimization of function~\eqref{eq:ObjectiveCorrentropy} can be expressed as
%\begin{equation*}
%	\min_{\bX} -\sum_{l=1}^L \langle \Phi(\by_{l*}) ,  \Phi(\widehat{\by}_{l*}) \rangle_\cp{H},
%\end{equation*}
%where $\langle \Phi(\cdot), \Phi(\cdot) \rangle_\cp{H}$ denotes the inner product in $\cp{H}$.
%}
%%The correntropy induces a newmetric that, as the distance
%%between and gets larger, the equivalent distance evolves
%%from 2-norm to 1-norm and eventually to zero-norm when
%%and are far apart
%

\section{ADMM for Solving the Correntropy-based Unmixing Problems}\label{sec:Algorithm}

We first briefly review the alternating direction method of multipliers (ADMM), following the expressions in~\cite[Chap.~3]{boyd2011distributed}. Consider %Assuming that we want to solve a
an optimization problem of the form
\begin{align*}
\displaystyle \min_\bx~~& f(\bx)+g(\bx),
\end{align*}
where the functions $f$ and $g$ are closed, proper and convex. The ADMM solves the equivalent constrained problem% of the form
\begin{align}\label{eq:ADMM}
\begin{aligned}
\displaystyle \min_{\bx,\bz} f(\bx)+g(\bz) %\\
\text{ subject to } {\bA}\bx+\bB\bz=\bc,
\end{aligned}
\end{align}
%with variables $\bx\in\mathds{R}^{n}$ and $\bz\in\mathds{R}^{m}$, where $A\in \mathds{R}^{p\times n}$ and $B\in \mathds{R}^{p\times m}$ and $\bc\in \mathds{R}^{p}$.
such as having the particular constraint $\bx=\bz$ for instance. While this formulation may seem trivial, the optimization problem can now be tackled using the augmented Lagrangian method where the objective function is separable in $\bx$ and $\bz$. By alternating on each variable separately, the ADMM repeats a direct update of the dual variable. In its scaled form, the ADMM algorithm is summarized in Algorithm~\ref{alg:ADMM}.
\begin{algorithm}[h]
\caption{The %Alternating Direction Method of Multipliers (
ADMM algorithm~\cite{boyd2011distributed}}
\label{alg:ADMM}
\begin{algorithmic}[1]
\REQUIRE
functions $f$ and $g$, matrices $\bA$ and $\bB$, vector $\bc$, parameter $\rho$
\STATE Initialize $k=0$, $\bx_0$, $\bz_0$ and $\bu_0$
\REPEAT
\STATE {$\bx_{k+1}=\arg\min_\bx f(\bx)+ \frac{\rho}{2}\|\bA\bx+\bB\bz_k-\bc+\bu_k\|_2^2$};
\STATE {$\bz_{k+1}=\arg\min_\bz g(\bz)+ \frac{\rho}{2}\|\bA\bx_{k+1}+\bB\bz-\bc+\bu_k\|_2^2$};
\STATE {$\bu_{k+1}=\bu_k+\bA\bx_{k+1}+\bB\bz_{k+1}-\bc$};
\STATE {$k=k+1$};
\UNTIL {stopping criterion}
\end{algorithmic}
\end{algorithm}

\subsection{Correntropy-based unmixing with full-constraints}\label{subsection: CUSAL_FC}

In the following, we apply the ADMM algorithm to solve the correntropy-based unmixing problem in the fully-constrained case, presented in \eqref{eq:SumToOne}. The main steps are summarized in Algorithm~\ref{alg:SumToOne}.
Rewrite the variables to be optimized in a vector $\bx\in \mathds{R}^{RT\times 1}$, which is stacked by the columns of the matrix $\bX$, namely $\bx = [ \bx_1^{\top} ~ \cdots ~ \bx_T^{\top} ]^\top$.
Rewrite also the following vectors in $\mathds{R}^{RT\times 1}$: 
%\begin{align*}
${\bz} = [ \bz_1^{\top} ~ \cdots ~ \bz_T^{\top} ]^\top$ 
and %\\
${\bu} = [ \bu_1^{\top} ~ \cdots ~ \bu_T^{\top} ]^\top$,
%\end{align*}
% in matrices $\bZ= [ \bz_1 ~ \cdots ~ \bz_T ] \in\mathds{R}^{R\times T}$ and $\bU=[ \bu_1 ~ \cdots ~ \bu_T ] \in\mathds{R}^{R\times T}$
where, for $t=1, \ldots, T$, ${\bz_t} = [z_{1t} ~ \cdots ~ z_{Rt}]^\top$ and $
{\bu_t} = [u_{1t} ~ \cdots ~ u_{Rt}]^\top$. By following the formulation of the ADMM in Algorithm~\ref{alg:ADMM}, we set
\begin{align}
\label{eq:f}
f(\bx)&=\mathcal{C}(\bx)+ \sum_{t=1}^T \iota_{{\{1\}}}(\cb1^{\top}\bx_t)\\
 g(\bz)&=\iota_{\mathds{R}_+^{RT}}(\bz) \nonumber\\
 {\bA} &= -\cb{I}, \bB= \cb{I}~~\text{and}~~\bc = \cb0, \nonumber
\end{align}
where $\cb{I}$ is the identity matrix, $\cb0 \in\mathds{R}^{RT\times 1}$ is the zero vector and $\iota_{\mathcal{S}}(u)$ is the indicator function of the set $\mathcal{S}$ defined by
\begin{equation*}
	\iota_{\mathcal{S}}(u) =\left\{\begin{array}{ll}
	 0 \qquad~\,\text{if}~u \in \mathcal{S}; \\
     \infty \qquad\text{otherwise}.%~\bx \notin \mathcal{S}.
	\end{array}\right.
\end{equation*}

In this case, the subproblem of the $\bx$-update (in line 3 of Algorithm~\ref{alg:ADMM}) addresses a nonconvex problem without any closed-form solution. To overcome this difficulty, we apply an inexact ADMM variant in lines 3-5 of Algorithm~\ref{alg:SumToOne}, which solves the subproblem iteratively using the gradient descent method, instead of solving it exactly and explicitly.

Before that, we eliminate the $T$ equality constraints, {\em i.e.}, the sum-to-one constraints, by replacing $x_{Rt}$ with
$$x_{Rt} = 1 - \sum_{r=1}^{R-1} x_{rt},$$
for $t =1, \ldots, T$. Let $\overline{\bx}\in \mathds{R}^{(R-1)T\times1}$ be the reduced vector of $(R-1)$ unknowns to be estimated, stacked by
$$\overline{\bx_t}=\left[ x_{1t} ~ \cdots ~ x_{(R-1)t} \right]^\top,$$ for $t =1, \ldots, T$.
By this means, the objective function in~\eqref{eq:f} is transformed from \eqref{eq:ObjectiveCorrentropy2} into the reduced-form%~\eqref{eq:Reducedf} (given on the top of next page),
\begin{equation}\label{eq:Reducedf}
f_1(\overline{\bx})= - \sum_{l=1}^{L} \exp\left(\frac {-1}{2\sigma^2}\sum_{t=1}^T \epsilon_l(\overline{\bx_t})^2 \right),
\end{equation}
where
%\begin{equation*}
 $\epsilon_l(\overline{\bx_t})=y_{lt} - m_{lR}- \sum_{p=1}^{R-1} (m_{lp}-m_{lR}) x_{pt}$,
%\end{equation*}
for $l =1, \ldots, L$. 
The gradient of~\eqref{eq:Reducedf} with respect to $\overline{\bx}$ is stacked as
$$\frac{\partial f_1}{\partial \overline{\bx}} = \left[ \frac{\partial f_1}{\partial \overline{\bx}_1}^\top ~ \cdots ~ \frac{\partial f_1}{\partial \overline{\bx}_T}^\top \right]^\top \in\mathds{R}^{ (R-1)T\times1},$$
where $\frac{\partial f_1}{\partial \overline{\bx}_t} = \left[ \frac{\partial f_1}{\partial \overline{x}_{1t}} %~~ \frac{\partial f_1}{\partial \overline{x}_{2t}}
~ \cdots ~\frac{\partial f_1}{\partial \overline{x}_{(R-1)t}} \right]^\top$, with the entries given by
\begin{equation*}%\label{eq:gradient}
\frac{\partial f_1(\overline{\bx})}{\partial \overline{x_{rt}}} \!=\! \tfrac{1}{\sigma^2} \!\!\sum_{l=1}^L (m_{lR}-m_{lr})\exp\!\Big(\tfrac {-1}{2\sigma^2}\sum_{s=1}^T \epsilon_l(\overline{\bx}_s)^2\Big) \epsilon_l(\overline{\bx_t}),
\end{equation*}
for all $r =1, \ldots, (R-1)$ and $t =1, \ldots, T$.
Similarly, the function $\frac{\rho}{2}\|{\bx}-\bz_k-{\bu_k}\|_2^2$ is expressed with respect to $\overline{\bx}$ as 
\begin{equation*}
\phi(\overline{\bx}) \!=\! \frac{\rho}{2} \sum_{t=1}^{T} \Big(\!1-\!\sum_{p=1}^{R-1}x_{pt}- z_{Rt,{k}} -u_{Rt,{k}} \Big)^2 \!+ \sum_{p=1}^{R-1} \left(x_{pt}- z_{pt,{k}}-u_{pt,{k}}\right)^2
\end{equation*}
with the entries in its gradient $\frac{\partial \phi}{\partial \overline{\bx}}$ given by
\begin{equation}\label{eq:gradientPhi}
\frac{\partial \phi(\overline{\bx})}{\partial \overline{x_{rt}}}= \rho \Big(x_{rt}+\sum_{p=1}^{R-1}x_{pt}-1+z_{Rt,k}-z_{rt,k}+u_{Rt,k}-u_{rt,k} \Big),
\end{equation}
for all $r =1, \ldots, R-1$ and $t =1, \ldots, T$.

The solution of the $\bz$-update in line 4 Algorithm~\ref{alg:ADMM} becomes the projection of $\bx_{k+1}-\bu_{k}$ onto the first orthant, as shown in line 7 of Algorithm~\ref{alg:SumToOne}.

\begin{algorithm}[h]
\caption{Correntropy-based unmixing with full-constraints (\textbf{CUSAL-FC})}
\label{alg:SumToOne}
\begin{algorithmic}[1]
%\REQUIRE
%functions $f$ and $g$, matrices $\bA$ and $\bB$, vector $\bc$, parameter $\rho$ and step-size $\eta$
\STATE{Initialize $k=0$, $\rho>0$, $\eta>0$, $\sigma>0$%, norms of the primal and dual residuals $r_{0}=%10^5$ and $
%s_{0}=10^5$
;  $\bx_0$, $\bz_0$ and $\bu_0$; }
\REPEAT
\REPEAT
\STATE {$\overline{\bx}_{k+1}=\overline{\bx}_{k+1}- \eta (\frac{\partial f_1}{\partial \overline{\bx}_{k+1}}+\frac{\partial \phi}{\partial \overline{\bx}_{k+1}})$};
\UNTIL{convergence}
\STATE {reform $\bx_{k+1}$ using $\overline{\bx}_{k+1}$};
\STATE {$\bz_{k+1}=\max(\cb0, \bx_{k+1}-\bu_{k})$};
\STATE {$\bu_{k+1}=\bu_k-(\bx_{k+1}-\bz_{k+1})$};
%\STATE {$r_{k+1}=\|\bx_{k+1}-\bz_{k+1}\|_2$, $s_{k+1}=\rho \|\bz_{k+1}-\bz_{k}\|_2$};
\STATE {$k=k+1$};
\UNTIL{stopping criterion}
%\ENSURE {$\bx$ and }
\end{algorithmic}
\end{algorithm}

\subsection{Sparsity-promoting  unmixing algorithm}\label{subsection: CUSAL_SP}

In order to apply the ADMM algorithm, we express the constrained optimization problem~\eqref{eq:Sparsity} as follows
\begin{align}\label{eq:sparsityADMM}
f(\bx)&=\mathcal{C}(\bx)\\
g(\bx)&=\iota_{\mathds{R}_+^{RT}}(\bx)+ \lambda \|\bx\|_1 \nonumber\\
 {\bA} &= -\cb{I}, \bB= \cb{I}~~\text{and}~~\bc = \cb0. \nonumber
\end{align}
By analogy with the previous case, the $\bx$-update in line 3 of Algorithm~\ref{alg:ADMM} is solved iteratively with the gradient descent method and is given in Algorithm~\ref{alg:Sparsity} lines 3-5.
The gradient of~\eqref{eq:sparsityADMM} with respect to ${\bx}$%, namely $\frac{\partial f}{\partial {\bx}}\in\mathds{R}^{ RT\times1}$,
 is stacked by $\frac{\partial f}{\partial {\bx}_t}$, where
%{\red Il faut mettre $y_{lt}$ et $y_{ls}$ pas en bold (et aussi les autres \'equations et pour $x_rt$). Autre chose, la notation {\em \`a la matlab} $.*$ ne doit pas \^etre \'ecrite dans un article.}
\begin{equation*}%\label{eq:gradient}
%\frac{\partial f}{\partial \bx_{t}}=-\frac{1}{\sigma^2} \sum_{l=1}^L(y_{lt} - \bm_{l}^{\top}\bx_{t}) \exp\left(\frac {-1}{2\sigma^2} \sum_{s=1}^T (y_{ls} -  \bm_{l}^{\top}\bx_{s})^2\right)\centerdot~\bm_{l}^{\top},
\frac{\partial f}{\partial \bx_{t}}=-\frac{1}{\sigma^2} \sum_{l=1}^L \epsilon_l(\bx_t) \exp\left(\frac {-1}{2\sigma^2} \sum_{s=1}^T (\epsilon_l(\bx_s))^2\right)~\bm_{l}^{\top},
\end{equation*}
for $t =1, \ldots, T$, where $\epsilon_l(\bx_t) = y_{lt} - \sum_{r=1}^R x_{rt} \, m_{lr}$.

The $\bz$-update in line 4 Algorithm~\ref{alg:ADMM} involves solving
\begin{equation}\label{eq:z-updateSparse}
  \bz_{k+1}=\arg\min_\bz \iota_{\mathds{R}_+^{RT}}(\bz)+ ({\lambda}/{\rho}) \|\bz\|_1+ \frac{1}{2}\|\bz-\bx_{k+1}-\bu_{k}\|_2^2.
\end{equation}
In \cite{boyd2011distributed}, the ADMM has been applied to solve various $\ell_1$-norm problems, including the well-known LASSO \cite{tibshirani1996regression}. The only difference between~\eqref{eq:z-updateSparse} and the $\bz$-update in LASSO is that in the latter, no non-negativity term $\iota_{\mathds{R}+^{RT}}(\bz)$ is enforced. In this case, the $\bz$-update in LASSO is the element-wise soft thresholding operation
\begin{equation*}
 \bz_{k+1}=S_{\lambda/\rho}(\bx_{k+1}-\bu_{k}),
\end{equation*}
where the soft thresholding operator \cite{boyd2011distributed} is defined by
\begin{equation*}
S_{b}(\zeta )= \left\{\begin{array}{@{}c@{\;}l}
 \zeta-b&~~~ \text{ if } \zeta>b; \\
 0&~~~ \text{ if } \|\zeta\|<b;  \\
 \zeta+b&~~~\text{ if } \zeta<-b.
	\end{array}\right.
\end{equation*}
%$S_{b}(\zeta )=(\zeta-b)_+-(-\zeta-b)_+$, where $(\dot)_+$ and $(\dot)_-$ are
Following~\cite{bioucas2010alternating}, it is straightforward to project the result onto the nonnegative orthant in order to include the non-negativity constraint, thus yielding
\begin{equation*}
  \bz_{k+1}=\max(\cb0, S_{\lambda/\rho}(\bx_{k+1}-\bu_{k})),
\end{equation*}
where the maximum function is element-wise. All these results lead to the correntropy-based unmixing algorithm with sparsity-promoting, as summarized in Algorithm~\ref{alg:Sparsity}.
 %the basis pursuit \cite{chen1998atomic}.
\begin{algorithm}[h]
\caption{Correntropy-based unmixing with sparsity-promoting (\textbf{CUSAL-SP})) }
\label{alg:Sparsity}
\begin{algorithmic}[1]
%\REQUIRE
%functions $f$ and $g$, matrices $\bA$ and $\bB$, vector $\bc$, parameter $\rho$, step-size $\eta$ and sparseness parameter $\lambda$
\STATE{Initialize $k=0$, $\rho>0$, $\sigma>0$, $\eta>0$, $\lambda>0$; $\bx_0$, $\bz_0$ and $\bu_0$;}
\REPEAT
\REPEAT
\STATE {${\bx}_{k+1}={\bx}_{k+1}- \eta \big(\frac{\partial f}{\partial{\bx}_{k+1}}+\rho(\bx_{k+1}-\bz_k-\bu_{k})\big)$  };
\UNTIL{convergence}
%\STATE {$\bv_k=\bx_{k+1}-\bu_{k}$};
\STATE {$\bz_{k+1}=\max(\cb0, S_{\lambda/\rho}(\bx_{k+1}-\bu_{k})
%\text{soft}(\bx_{k+1}-\bu_{k}, \frac{\lambda}{\rho})
)$};
\STATE {$\bu_{k+1}=\bu_k-(\bx_{k+1}-\bz_{k+1})$};
\STATE {$k=k+1$};
\UNTIL{stopping criterion}
\end{algorithmic}
\end{algorithm}

\subsection{On the initialisation and the bandwidth determination}

We apply a three-fold stopping criterion for Algorithms~\ref{alg:SumToOne} and~\ref{alg:Sparsity}, according to~\cite{boyd2011distributed, bioucas2010alternating}:
($\text{\romannumeral1}$) the primal and dual residuals are small enough, namely $\|\bx_{k+1}-\bz_{k+1}\|_2 \leq \epsilon_1$ and $\rho\|\bz_{k+1}-\bz_{k}\|_2 \leq \epsilon_2$, where $\epsilon_1=\epsilon_2= \sqrt{RT}\times 10^{-5}$ as in~\cite{bioucas2010alternating},
($\text{\romannumeral2}$) the primal residual starts to increase, {\em i.e.}, $\|\bx_{k+1}-\bz_{k+1}\|_2 > \|\bx_{k}-\bz_{k}\|_2$,
or ($\text{\romannumeral3}$) the maximum iteration number is attained.

The bandwidth $\sigma$ in the Gaussian kernel should be well-tuned.
%{\red Abde: Indeed, a high value for this parameter increases the robustness of the algorithm to outliers~\cite{he2011maximum}. However, this affects the algorithm convergence that requires a reduced value for $\sigma$.}
Note that a small value for this parameter punishes harder the outlier bands, thus increasing the robustness of the algorithm to outliers~\cite{he2011maximum}. %However, considering the finite number of samples (bands), a too small $\sigma$ yields meaningless estimation \cite{liu2007correntropy}, and also affects the algorithm convergence that requires a relatively high value for this parameter.
%On one hand, it is an essential parameter which controls the robustness of the correntropy to outliers~\cite{he2011maximum}. On the other hand, the convergence of the algorithm should achieve with it.
Note that, in this study, the ADMM is applied to address a nonconvex objective function, thus no convergence is guaranteed theoretically, according to~\cite{boyd2011distributed}. Considering these issues, we propose to fix the bandwidth empirically as summarized in Algorithm~\ref{alg:BandwidthTuning} and described next.

Following~\cite{he2011maximum, WangCorrentropy}, we first initialize the bandwidth parameter as a function of the reconstruction error, given by
\begin{equation}\label{eq:Sigma}
 \sigma_0^2= \frac{R}{8L}\|\bY-\bM \bX_\text{LS}\|_F^2, % Initialiser Sigma, increase it with Sigma=Sigma*1.2 until work
\end{equation}
where $\bX_\text{LS}$ is the least-squares solution \eqref{eq:ls_solution}. In the case of %meaningless estimation, namely the
a result too apart from that of least-squares solution, the parameter is augmented by $\sigma= 1.2\sigma$, until that the condition
%\begin{equation*}
$\frac{\|\bY-\bM\bX\|_F}{\|\bY-\bM\bX_\text{LS}\|_F}<2$
%\end{equation*}
is satisfied. The algorithm divergence occurs if the stopping criterion (ii) is satisfied, namely the primal residual increases over iterations. In this case, either the parameter is too large due to an overestimated initialization, or it is too small. Accordingly, we either decrease it by $\sigma=\sigma_0/p$, or increase it by $\sigma=1.2\sigma$, until that the ADMM converges. %See Algorithm~\ref{alg:BandwidthTuning} for more details.

\begin{algorithm}[h]
\caption{Tuning the bandwidth parameter $\sigma$}
\label{alg:BandwidthTuning}
\begin{algorithmic}[1]
\STATE{Initialize $\sigma=\sigma_0$ using \eqref{eq:Sigma}; $p=1$;}
\STATE{Do CUSAL with Algorithm~\ref{alg:SumToOne} or Algorithm~\ref{alg:Sparsity}};
\IF{stopping criterion (i) or (iii) is satisfied }
   \IF{condition $\frac{\|\bY-\bM\bX\|_2}{\|\bY-\bM\bX_\text{LS}\|_2}<2$ is satisfied,}
   \STATE{$\sigma^* = \sigma$ (optimal value)}
   \ELSE
   \STATE %{meaningless estimation occurs,}\\
   {increase $\sigma =1.2 \sigma$, and go to line 2}
   \ENDIF
\ELSE  %{no convergence }
   \IF{$\sigma > 1000 \sigma_0$ (due to the overestimated $\sigma_0$) }
   \STATE{$p=p+1$;}
   \STATE{decrease $\sigma = \sigma_0/p$, and go to line 2}
%   \ELSIF{$p\geq 50$}
%   \STATE{break}
   \ELSE
   \STATE{increase $\sigma =1.2 \sigma$, and go to line 2}
   \ENDIF
\ENDIF
\end{algorithmic}
\end{algorithm}

\section{Experiments with Synthetic Data}\label{sec: Synthetic}

In this section, the performance of the proposed fully-constrained (CUSAL-FC) and sparsity-promoting (CUSAL-SP) algorithms is evaluated on synthetic data. A comparative study is performed considering six state-of-the-art methods proposed for linear and nonlinear unmixing models.
%In this section, the performance of the proposed CUSAL algorithms, in both the fully-constrained (CUSAL-FC) and sparsity-promoting (CUSAL-SP) versions, are evaluated on synthetic data. A comparative study is performed considering six state-of-the-art methods proposed for linear and nonlinear unmixing models.
%The proposed CUSAL algorithms are compared to the following methods:
\begin{itemize}
\item
{Fully-Constrained Least-Squares} (\textbf{FCLS})~\cite{Heinz}: The FCLS is developed for the linear model. Enforcing both ANC and ASC constraints, this technique yields the optimal abundance matrix in the least-squares sense.
\item
{Sparse Unmixing by variable Splitting and Augmented Lagrangian} (\textbf{SUnSAL})~\cite{bioucas2010alternating}: This method is based on the ADMM. %Under the linear assumption, 
Several variants are developed by including different constraints, with the fully-constrained SUnSAL-FCLS and the sparsity-promoting SUnSAL-sparse. %the alternating direction method of multipliers
\item
{The Bayesian algorithm for Generalized Bilinear Model }(\textbf{BayGBM})~\cite{HaAlDoTo2011,halimi2011unmixing}:
This method estimates the abundances with the generalized bilinear model (GBM), which adds second-order interactions between endmembers to the linear model, yielding the model
\begin{equation*}%\label{eq:GBM}
  \by_{t}=\bM\bx_t + \sum_{i=1}^{R-1}\sum_{j=i+1}^{R} \gamma_{ij,t} x_{it}x_{jt}(\bm_i \odot\bm_j)+\bn_t,
\end{equation*}
where $0 \leq \gamma_{ij,t}\leq 1$ controls the interactions between endmembers $\bm_i$ and $\bm_j$%, $\bn_t \in \mathds{R}^{L}$ represents the additive Gaussian noise
, and $\odot$ is the element-wise product. The BayGBM considers both ANC and ASC. %Appropriate prior distributions are chosen for the parameters under estimation, then the joint posterior distributions are derived. The unknown parameters are estimated by the gradient-based method.
\item
{The Bayesian algorithm for Polynomial Post-Nonlinear Mixing Model} (\textbf{BayPPNMM})~\cite{altmann2012supervised}:
This algorithm estimates the parameters by %involved in the polynomial post-nonlinear mixing model (PPNMM), which 
assuming that the pixel reflectances are nonlinear functions of endmembers using% the model
\begin{equation}\label{eq:PPNMM}
  \by_{t}=\bM\bx_t + b_t (\bM\bx_t) \odot (\bM\bx_t)+\bn_t,
\end{equation}
where the nonlinear terms are characterized by $b_t \in \mathds{R}$% compared with GBM model~\eqref{eq:GBM}
, and both ANC and ASC are required.
\item
{Kernel Fully-Constrained Least-Squares}
(\textbf{KFCLS})~\cite{broadwater2007kernel}: This method generalizes FCLS, by replacing the inner product with a kernel function.
 %For the case with linear kernel, KFCLS reduces to FCLS.
In the following, the Gaussian kernel is applied for simulation. %There are two parameters to adjust, the kernel bandwidth $\sigma$, and the relaxation parameter $\nu$ controlling the sum-to-one constraint. To this end, a
%A preliminary test on 100 randomly selected pixels is performed for each data under study, using the set $\{1, 2, 3, 4, 5\}$ for the parameter $\sigma$ and the set $\{1, 10^{-1}, 10^{-2}, 10^{-3}, 10^{-4}\}$ for the parameter that controls the relaxation of the sum-to-one constraint.%$\nu$.
\item
{Robust nonnegative matrix factorization}
(\textbf{rNMF})~\cite{Fevotte15}: To capture the nonlinear effect (outliers), this NMF-based method introduces a group-sparse regularization term into the linear model. Accounting for both constraints, the problem is optimized by a block-coordinate descent strategy. For fair comparison in this paper, the endmembers are fixed with the real ones. The regularization parameter is set with the degree of sparsity as suggested in ~\cite{altmann2012supervised}.%, as described in this paper.
\end{itemize}
%In the following, the initial value of $\sigma$ is set by~\eqref{eq:Sigma} with $\beta=\frac{1}{2}$.
%\subsection{Performance of CUSAL-FC (fully-constrained unmixing algorithm)}
%\begin{figure}
%\centering
%\graphicspath{{Graphics/}}
%\centering
%\includegraphics[trim = 0mm 0mm 0mm 0mm, clip,width=0.45\textwidth]{EndmembersLib.eps}
%\caption{The endmembers selected from the USGS library.}
%\label{Fig. EndmembersLib}
%\end{figure}
% K-Hype~\cite{13.tsp.unmix} and GBM-Bayes \cite{HaAlDoTo2011}.

%\subsection{Performance of CUSAL-FC (fully-constrained algorithm)}

We first compare the fully-constrained CUSAL-FC, presented in~\ref{subsection: CUSAL_FC}, with the state-of-the-art methods. A series of experiments are performed, mainly considering the influences of four aspects: ($\text{\romannumeral1}$) mixture model, ($\text{\romannumeral2}$) noise level, ($\text{\romannumeral3}$) number of corrupted bands and ($\text{\romannumeral4}$) number of endmembers.
%We consider 19 spectra selected from the United States Geological Survey (USGS) digital spectral library~\cite{Dias08}, as illustrated in ~\figurename~\ref{Fig. EndmembersLib}. These spectra, namely the endmembers, are measured over $L=244$ continuous bands with the wavelength ranging from 0.2$\mu m$ to 3.0$\mu m$.

\begin{figure}
\centering
\small
\graphicspath{{Graphics/Synthetic/}}
\psfrag{Bands}{\small Bands}
\psfrag{Reflectance}{\small Reflectance}
\includegraphics[trim = 6mm 0mm 11mm 5mm, clip,width=.45\textwidth]{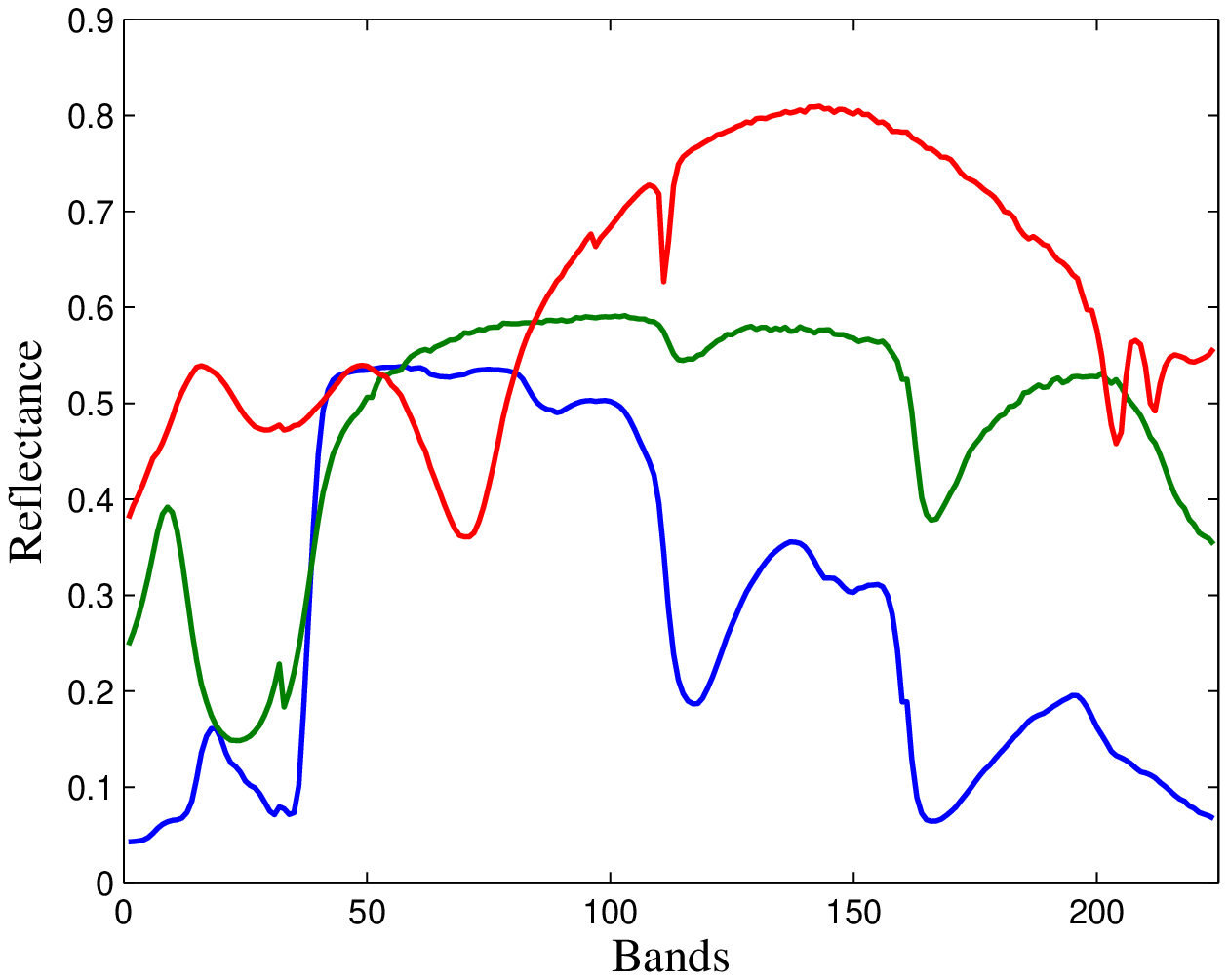}
\includegraphics[trim = 6mm 0mm 11mm 5mm, clip,width=.45\textwidth]{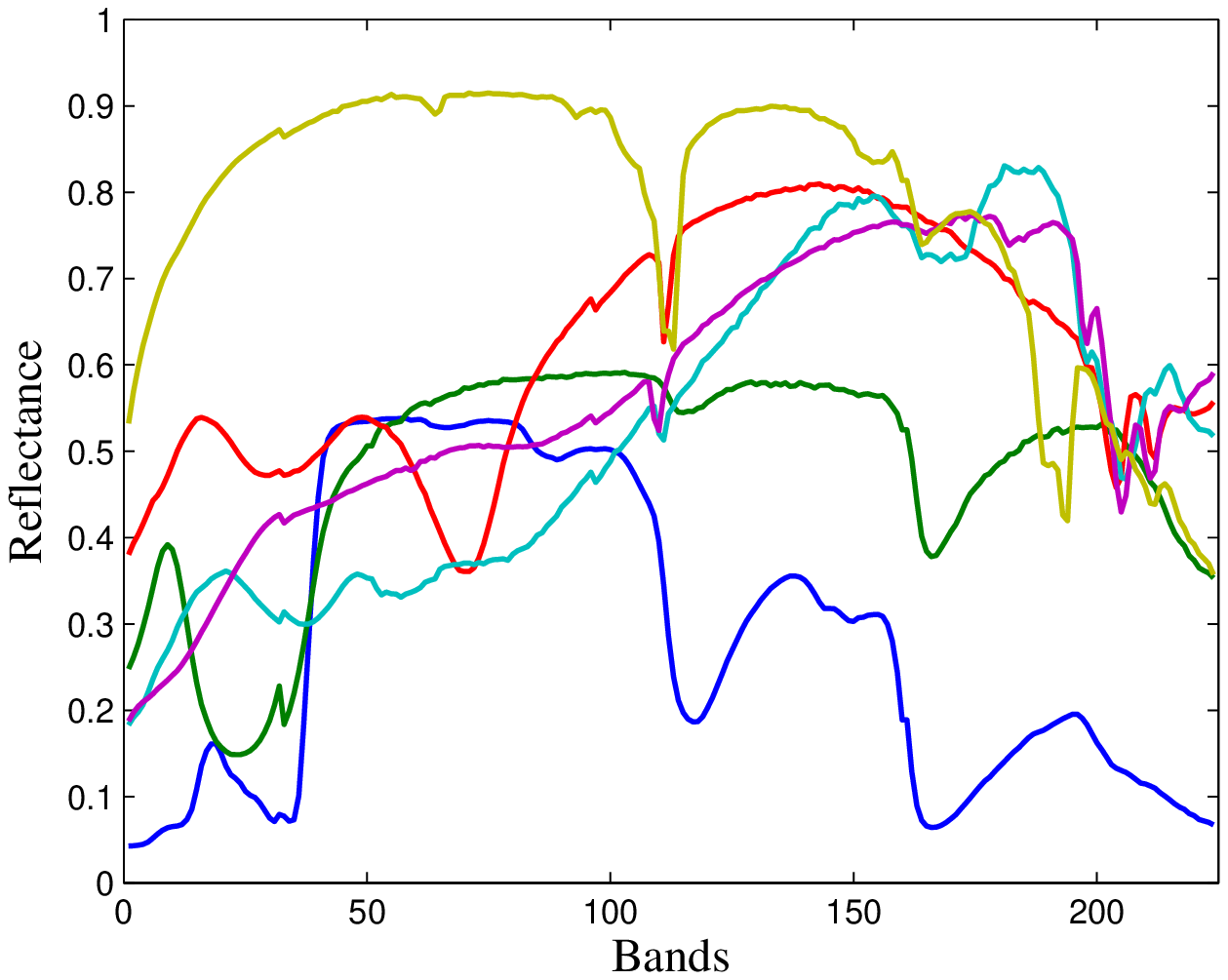}
\caption{The $R=3$ (left) and $6$ (right) USGS signatures chosen for simulation.}
\label{Fig. Endmember}
\end{figure}

Each image, of $50 \times 50$ pixels, is generated using either the linear mixing model~\eqref{eq:LMM} or the polynomial post-nonlinear mixing model (PPNMM)~\eqref{eq:PPNMM}, where $\bn_t$ is a Gaussian noise of $\textrm{SNR}\in \{15, 35\}\textrm{dB}$.
The $R\in \{3, 6\}$ endmembers, as shown in~\figurename~\ref{Fig. Endmember}, are drawn from %20 spectra selected from %
the USGS digital spectral library~\cite{Dias08}. These endmembers are defined over $L=244$ continuous bands with the wavelength ranging from 0.2$\mu m$ to 3.0$\mu m$. The abundance vectors $\bx_t$ are uniformly generated using a Dirichlet distribution as in~\cite{Dias08, Halimi15}. For PPNMM, the values of $b_t$ are generated uniformly in the set $(-3, 3)$ according to \cite{altmann2012supervised}.
To imitate the noisy bands in the real hyperspectral images, several bands in the generated data are corrupted by replacing the corresponding rows of $\bY$ with random values within $[0, 1]$.
The number of corrupted bands varies in the set $\{0, 20, 40, 60\}$. %are firstly generated with a Gamma distribution with the shape parameter $k=1$ and the shape parameter $\theta=1$, and then normalized by $\|\bx_t\|_1$ to satisfy the "sum-to-one" constraint.

\begin{figure}
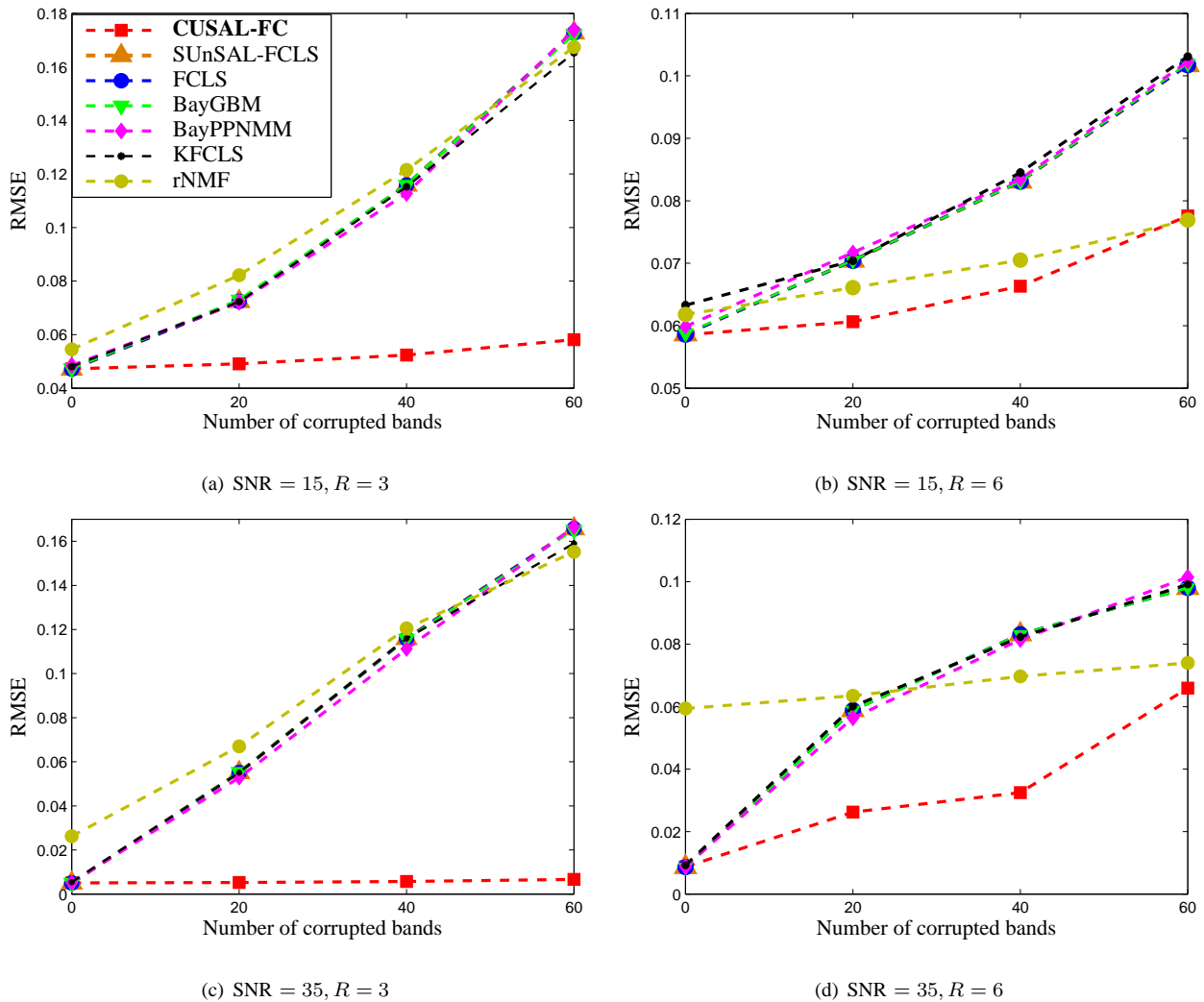

\centering
\graphicspath{{Graphics/Synthetic/CEU_FCLSVersionEndFixed/LMM_WithrLMM/}}
\centering
\small
 \psfrag{CUSAL-FC}{ \bf CUSAL-FC}
 \psfrag{SUnSAL-FCLS}{  SUnSAL-FCLS}
 \psfrag{FCLS}{  FCLS}
 \psfrag{BayGBM}{  BayGBM}
 \psfrag{BayPPNMM}{ BayPPNMM}
 \psfrag{KFCLS}{  KFCLS}
 \psfrag{rNMF}{  rNMF}
\subfigure[SNR $=15, R=3$]{
\includegraphics[trim = 3mm 0mm 10mm 5mm, clip,width=.45\textwidth]{SNR15_End3.eps}
}
\subfigure[SNR $=15, R=6$]{
\includegraphics[trim = 3mm 0mm 10mm 5mm, clip,width=.45\textwidth]{SNR15_End6.eps}
}
\subfigure[SNR $=35, R=3$]{
\includegraphics[trim = 3mm 0mm 10mm 5mm, clip,width=.45\textwidth]{SNR35_End3.eps}
}
\subfigure[SNR $=35, R=6$]{
\includegraphics[trim = 3mm 0mm 10mm 5mm, clip,width=.45\textwidth]{SNR35_End6.eps}
}
\caption{\textbf{LMM data: } The root mean square error (RMSE) with respect to the number of corrupted bands, averaged over ten Monte-Carlo realizations, for different number of endmembers %(left to right) 
and SNR%(top to bottom)
.}
%. \newline
%\textbf{Left to right: }%Comparison of the proposed CUSAL-FC, SUnSAL-FCLS, FCLS, BayGBM, BayPPNMM and KFCLS for
%Various number of endmembers. \textbf{Top to bottom:} %Comparison of the proposed CUSAL-FC, SUnSAL-FCLS, FCLS, BayGBM, BayPPNMM and KFCLS for various
%Different SNR.\newline}
%%\caption{\textbf{LMM data: }The averaged root mean square error (RMSE) with respect to the number of corrupted bands (10 Monte-Carlo realizations). \textbf{Left to right: }Comparison of the proposed CUSAL-FC, SUnSAL-FCLS, FCLS, BayGBM, BayPPNMM and KFCLS for various number of endmembers. \textbf{Top to bottom:} Comparison of the proposed CUSAL-FC, SUnSAL-FCLS, FCLS, BayGBM, BayPPNMM and KFCLS for various SNR.}
\label{Fig. SyntheticCUSALFCVersionLMM}
\end{figure}

%Each data is unmixed respectively by three comparing algorithms.
The unmixing performance is evaluated using the abundance root mean square error (RMSE)~\cite{HaAlDoTo2011, Yokoya14}, defined by
\begin{equation*}
   \mathrm{RMSE}=\sqrt{\frac{1}{RT}\sum_{t=1}^T \|\bx_{t}-\widehat{\bx}_{t}\|^2},\notag
\end{equation*}
where $\widehat{\bx}_{t}$ is the estimated abundance vector.
\figurename~\ref{Fig. SyntheticCUSALFCVersionLMM} and~\ref{Fig. SyntheticCUSALFCVersionPPNMM}
illustrates the average of RMSE over 10 Monte-Carlo realizations, respectively on the LMM and PPNMM data.
%As observed, independent of varying mixture models, noisy levels and numbers of endmembers under consideration, the proposed CUSAL-FC outperforms the comparing methods in terms of RMSE, when outlier bands exist in the data.
It is easy to see that, in presence of outlier bands, the proposed CUSAL-FC algorithm outperforms all the comparing methods in terms of RMSE, for different mixture models, noise levels and numbers of endmembers. It is also shown that the performance of the proposed algorithm improves when increasing the SNR.

\begin{figure}
\centering
\graphicspath{{Graphics/Synthetic/CEU_FCLSVersionEndFixed/PNMM_WithrLMM/}}
\centering
\small
 \psfrag{CEUnSAL-FC}{ \bf CUSAL-FC}
 \psfrag{SUnSAL-FCLS}{  SUnSAL-FCLS}
 \psfrag{FCLS}{  FCLS}
 \psfrag{BayGBM}{  BayGBM}
 \psfrag{BayPPNMM}{ BayPPNMM}
 \psfrag{KFCLS}{  KFCLS}
 \psfrag{rNMF}{  rNMF}
\subfigure[SNR $=15, R=3$]{
\includegraphics[trim = 3mm 0mm 10mm 5mm, clip,width=.45\textwidth]{SNR15_End3.eps}
}
\subfigure[SNR $=15, R=6$]{
\includegraphics[trim = 3mm 0mm 10mm 5mm, clip,width=.45\textwidth]{SNR15_End6.eps}
}
\subfigure[SNR $=35, R=3$]{
\includegraphics[trim = 3mm 0mm 10mm 5mm, clip,width=.45\textwidth]{SNR35_End3.eps}
}
\subfigure[SNR $=35, R=6$]{
\includegraphics[trim = 3mm 0mm 10mm 5mm, clip,width=.45\textwidth]{SNR35_End6.eps}
}
\caption{\textbf{PPNMM data: } The root mean square error (RMSE) with respect to the number of corrupted bands, averaged over ten Monte-Carlo realizations, for different number of endmembers %(left to right) 
and SNR% (top to bottom)
.}
%. \newline
%\textbf{Left to right: }Various number of endmembers. \textbf{Top to bottom:} %Comparison of the proposed CUSAL-FC, SUnSAL-FCLS, FCLS, BayGBM, BayPPNMM and KFCLS for various
%Different SNR.\newline}
%%Comparison of the proposed CUSAL-FC, SUnSAL-FCLS, FCLS, BayGBM, BayPPNMM and KFCLS  for various number of endmembers. \textbf{Top to bottom:} Comparison of the proposed CUSAL-FC, SUnSAL-FCLS, FCLS,  BayGBM, BayPPNMM and KFCLS for various SNR.}
\label{Fig. SyntheticCUSALFCVersionPPNMM}
\end{figure}

%\subsection{Performance of CUSAL-SP (sparsity-promoting algorithm)}
%
%This section compares the proposed sparsity-promoting CUSAL (CUSAL-SP) algorithm with the state-of-art methods. The algorithms are evaluated using synthetic data with sparse abundances when varying (i) the mixture model, (ii) the noise level, (iii) the  number of corrupted bands and (iv) the sparsity level of abundances.
The performance of the proposed the sparsity-promoting CUSAL-SP, presented in~\ref{subsection: CUSAL_SP}, is compared with the sparsity-promoting SUnSAL-sparse, as well as the FCLS, on a series of data with sparse abundance matrices. The influences of ($\text{\romannumeral1}$) the number of corrupted bands and ($\text{\romannumeral2}$) the sparsity level of the abundances, are studied. 
% and the Gaussian noise with $SNR=30$.
Each image, of $15 \times 15$ pixels, is generated by the linear mixture model. The endmember matrix is composed by $R=62$ USGS signatures, where the angle between any two different endmembers is larger than ${10^{\circ}}$~\cite{Iordache11}.
%Concerning the abundance matrix, a natural indicator of its sparsity level is the portion of zeros in it. To this end, a percentage of the entries in the abundance matrix, varying within the set $\{15$\%$, 30$\%$, 45$\%$, 60$\%, $75$\%$\}$, are set as zeros, while guaranteeing at least one nonzero entry in each column. %The non-null entries are then attributed by the uniform distribution within $[0,1]$, and finally normalized to satisfy the sum-to-one constraint.
The $K$ nonzero entries in each abundance vector $\bx_t$ are generated by a Dirichlet distribution. The value of $K$ ({\em i.e.}, the indicator of sparsity level) ranges from 4 to 20, while the number of corrupted bands varies in $\{0, 20, 40, 60\}$.
%%Let $K=\|\bx_t\|_0$,
%within a unit simplex of dimension $K-1$, where the $n$ unit simplex is defined as $\bigtriangleup^n= \{(t_1, t_2, \ldots, t_{n+1})| \sum_{i=1}^{n+1} t_i=1, t_i\geq0 \}$.
We set the Gaussian noise by $\textrm{SNR}=30\textrm{dB}$, a level that that is commonly present in real hyperspectral images according to~\cite{Iordache11}. For both sparsity-promoting algorithms, the regularization parameter $\lambda$ %needs to be tuned.
%A rough measure from the input spectra~\cite{Hoyer04} takes the form
%\begin{equation*}\label{eq:lambda}
%   s=\frac{1}{\sqrt{L}}\sum_{l=1}^L\frac{\sqrt{T}-{\|\by_{l*}\|_{1}}/{\|\by_{l*}\|_{2}}  }{\sqrt{T}-1}.
%\end{equation*}
%For both the CUSAL and SUnSAL,
%In the following simulations, it 
is adjusted %for the optimal performance, 
using the set %\lambda \in\{{10}^{-5}, 5\times {10}^{-5}, {10}^{-4}, 5\times {10}^{-4}, {10}^{-3}\}$ for CUSAL-SP, and the set 
$%\lambda \in
\{ {10}^{-5}, 5 \cdot {10}^{-5}, {10}^{-4}, 5\cdot {10}^{-4}, {10}^{-3}, {10}^{-2}, {10}^{-1}\}.$% for SUnSAL-Sparse,.

\begin{figure}
\centering
\graphicspath{{Graphics/Synthetic/CEU_SparseVersion/}}
\centering
\small
\psfrag{CUSAL-SP}{ \bf CUSAL-SP}
\psfrag{SUnSAL-sparse}{ SUnSAL-sparse}
\psfrag{FCLS}{ FCLS}
\psfrag{K}{ $K$}
\psfrag{SRE(dB)}{ SRE(dB)}
\subfigure[0 corrupted band]{
\includegraphics[trim = 5mm 0mm 10mm 5mm,  clip,width=.445\textwidth]{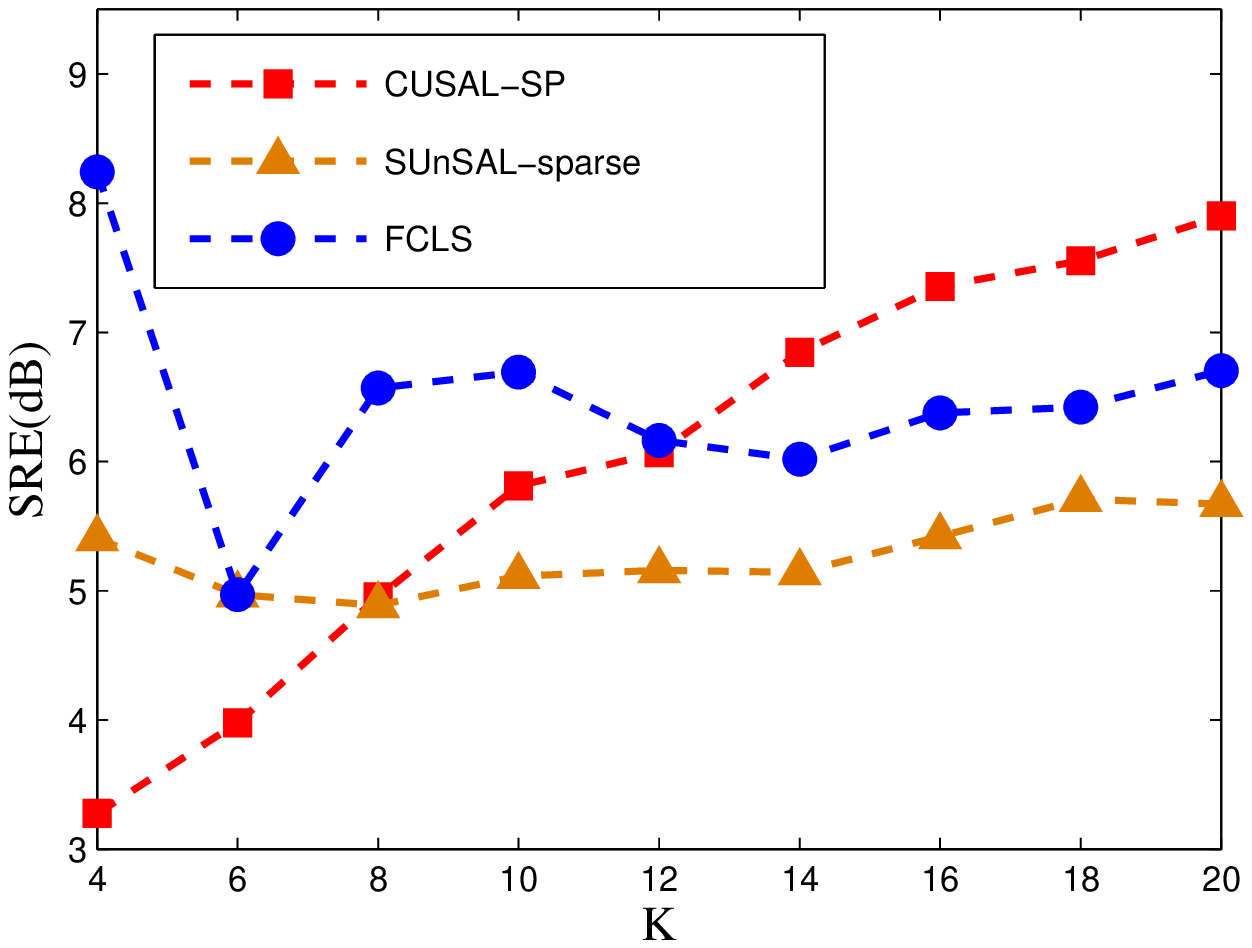}
}
\subfigure[20 corrupted bands]{
\includegraphics[trim = 5mm 0mm 10mm 5mm,  clip,width=.45\textwidth]{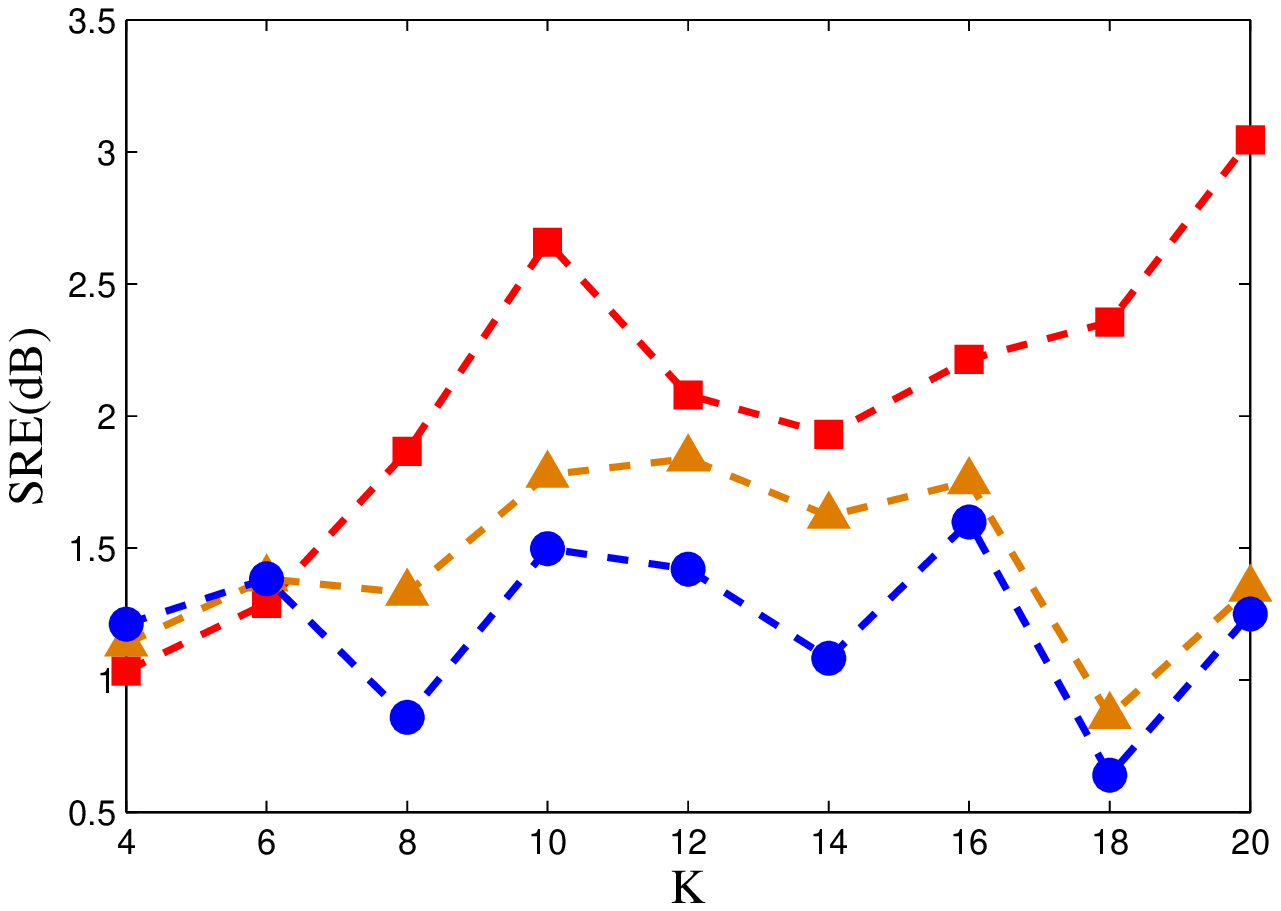}
}
\subfigure[40 corrupted bands]{
\includegraphics[trim = 5mm 0mm 10mm 5mm,  clip,width=.45\textwidth]{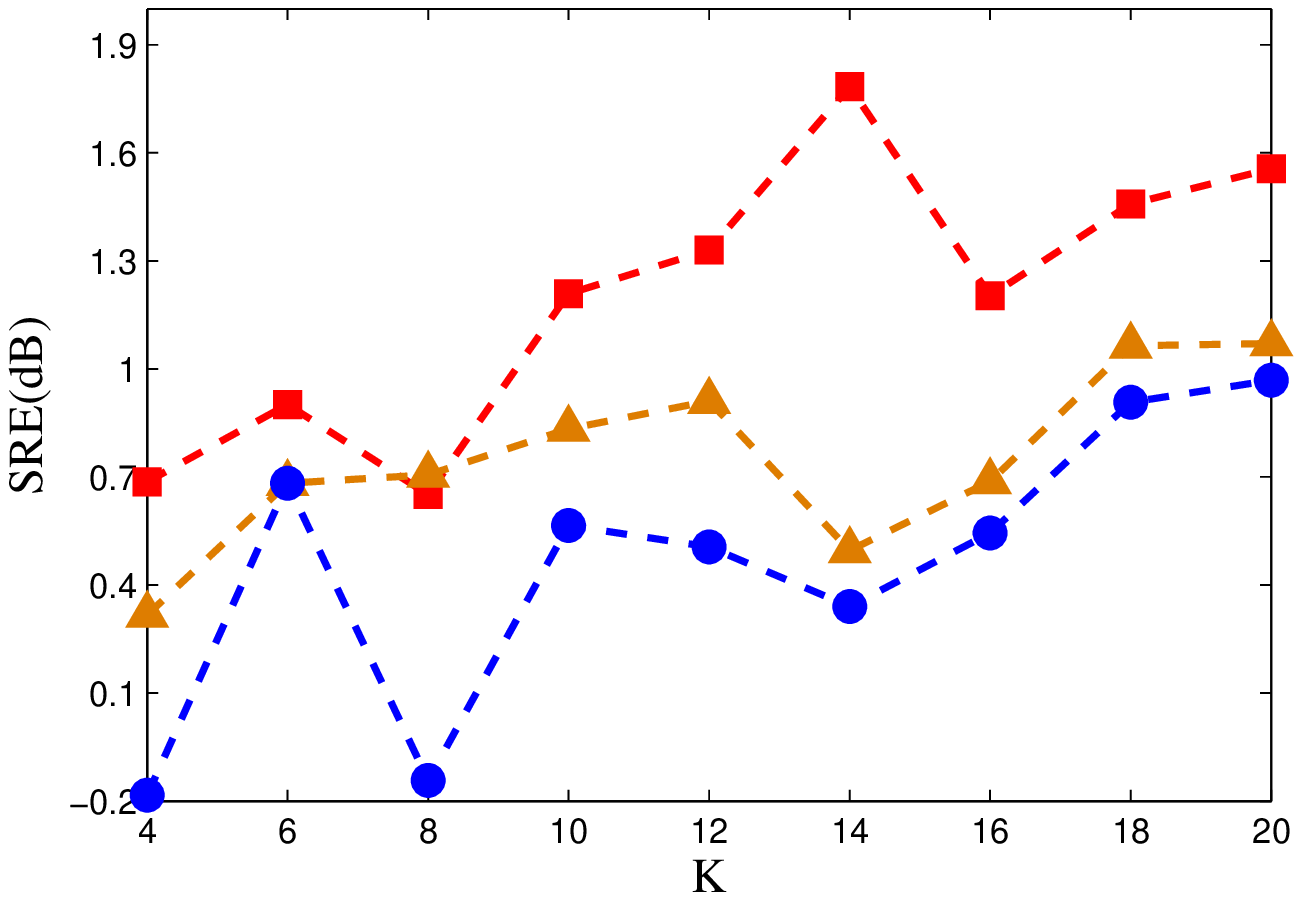}
}
\subfigure[60 corrupted bands]{
\includegraphics[trim = 5mm 0mm 10mm 5mm,  clip,width=.45\textwidth]{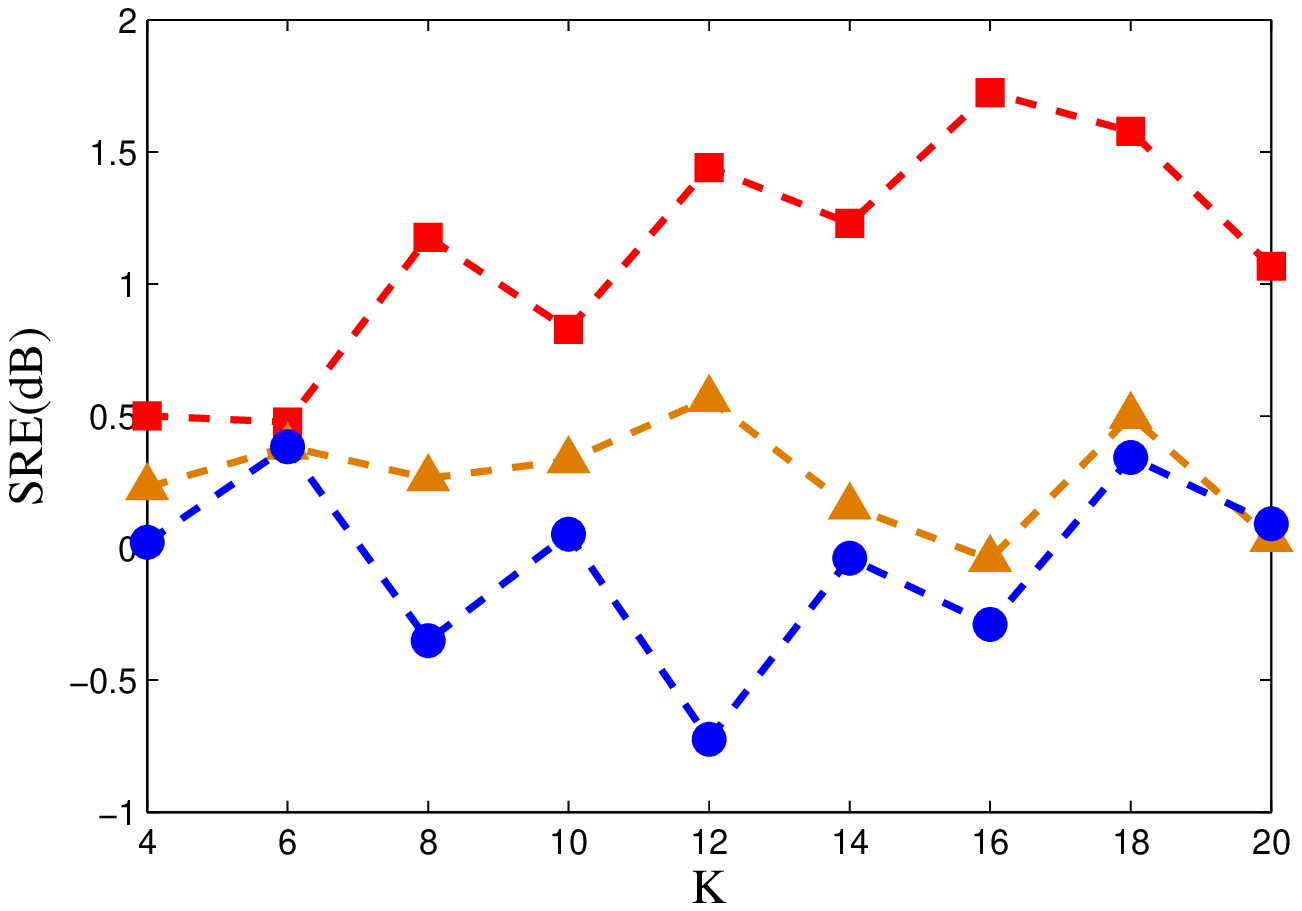}
}
\caption{\textbf{LMM data: } The averaged signal-to-reconstruction error (SRE) with respect to the sparsity level $K$, averaged over ten Monte-Carlo realizations. Comparison %of the proposed CUSAL-SP, SUnSAL-sparse and FCLS 
for various number of corrupted bands at SNR $=30$.}
\label{Fig. SyntheticCUSALSparseVersionLMM}
\end{figure}

The unmixing performance with the sparsity-promoting algorithms is evaluated using the signal-to-reconstruction error, measured in decibels, according to~\cite{bioucas2010alternating, Iordache11}. It is defined by
\begin{equation*}
   \mathrm{SRE}=10\log_{10} \Bigg( \frac{\sum_{t=1}^T\| \bx_t\|_2^2}{\sum_{t=1}^T \|\bx_t-\widehat{\bx}_t\|_2^2}\Bigg).
\end{equation*}
%where ${\bx_t}$ and $\widehat{\bx_t}$ are the real and estimated abundance vectors, respectively.
The results, averaged over ten Monte-Carlo realizations, are illustrated in \figurename~\ref{Fig. SyntheticCUSALSparseVersionLMM}. %We conclude that, independent of the varying noisy levels and numbers of endmembers under consideration, the proposed CUSAL outperforms SUnSAL and FCLS in terms of RMSE, when outliers exist in the data.
Considering that the abundance matrix under estimation is sparse at different levels, we conclude the following: 
%As the number of corrupted bands increases, the unmixing performance reduces, since the $\mathrm{SRE}$ generally decreases.
Concerning the case without outlier bands, the CUSAL-SP outperforms the SUnSAL-SAL for $K>8$ and FCLS for $K>12$. When the number of outlier bands is increases, the proposed CUSAL-SP algorithm generally provides the best unmixing quality with the highest SRE value, especially for $K>6$.   %This phenomenon is reasonable, since FCLS/ BayPPNMM estimates the abundance by assuming that the data is mixed by LMM/ PPNMM. %Another reason resides in the fact that the synthetic data is generated by imposing the ASC constraint, while for both the CUSAL and SUnSAL with the sparsity-promoting version, this constraint is relaxed.

\section{Experiments with Real Data}\label{sec: Real}

This section presents the performance of the proposed algorithms on a real hyperspectral image. We consider a $250\times190$ sub-image taken from the Cuprite mining image, acquired by the AVIRIS sensor when flying over Las Vegas, Nevada, USA. The image has been widely investigated in the literature \cite{13.tsp.unmix, Iordache11}.
The raw data contains $L=224$ bands, covering a wavelength range $0.4-2.5\mu m$. Among, there are $37$ relatively noisy ones with low SNR, namely the bands $1-3$, $105-115$, $150-170$, and $223-224$.
The geographic composition of this area is estimated to include up to $14$ minerals \cite{VCA}. Neglecting the similar signatures, we consider $12$ endmembers as often investigated in the literature \cite{13.tsp.unmix, lu2013manifold}.
The VCA technique is first applied to extract these endmembers on the clean image with $L=187$ bands. %The resulting $12$ pixel indices are retained.
%As opposed to the literatures that remove the noisy bands in prior,
Starting from $L=187$ bands, the noisy bands, randomly chosen from the bands $1-3$, $105-115$, $150-170$, and $223-224$, are gradually included to form a series of input data. Therefore, the experiments are conducted with $L=187, 193, 199, 205, 211, 217, 223$ and $224$ bands.

\begin{figure}[t]
\centering
 \graphicspath{{Graphics/Cuprite/}}
 \psfragscanon
 \psfrag{CUSAL-FC}{\bf CUSAL-FC}
 \psfrag{SAD}{ SAD}
 \psfrag{Number of bands}[c][c]{Number of bands}
\includegraphics[trim = 10mm 0mm 0mm 0mm, clip, width=.75\textwidth]{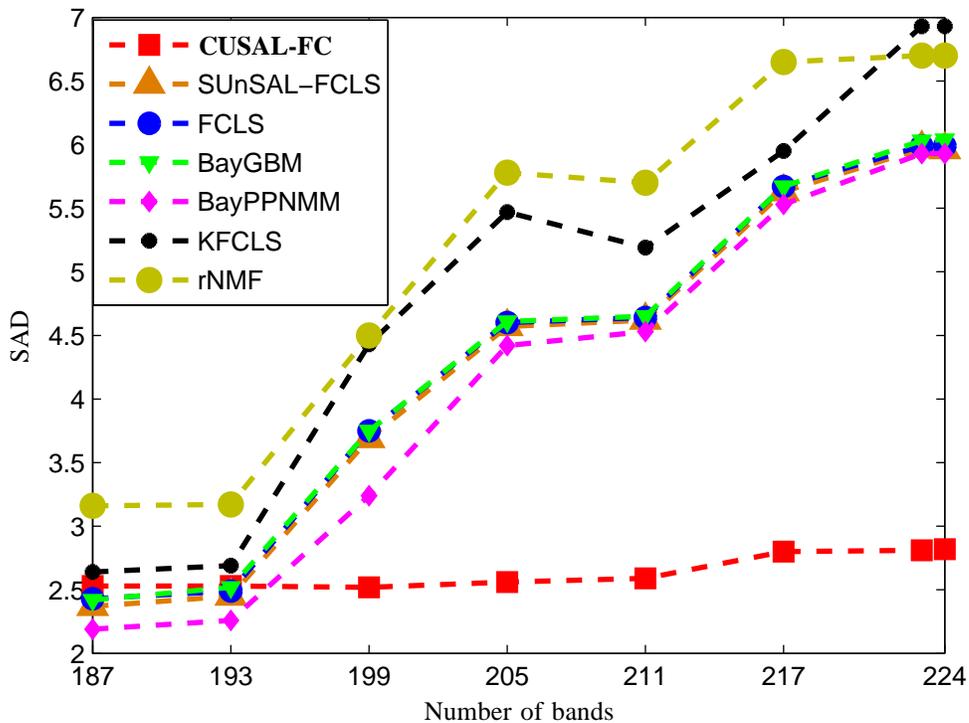}
\caption{\textbf{Cuprite image: } The averaged spectral angle distance (SAD) using different number of bands, computed without the noisy bands $1-3, 105-115, 150-170,$ and $223-224$.}
\label{Fig.SADCuprite}
\end{figure}

\begin{figure}
\centering
\graphicspath{{Graphics/Cuprite/}}
\centering
%\small
\includegraphics[trim = 39mm 13mm 9mm 73mm, clip,width=0.85\textwidth]{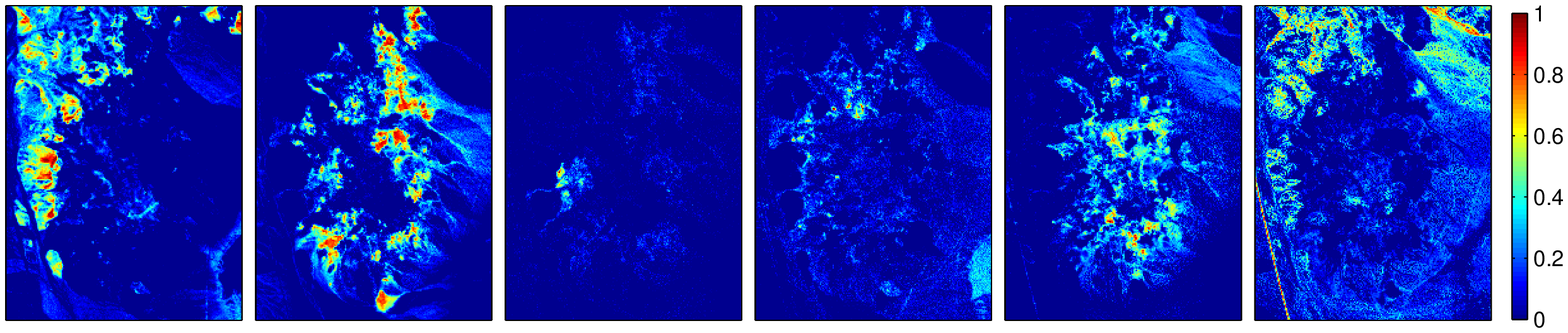}
\includegraphics[trim = 39mm 13mm 9mm 73mm, clip,width=0.85\textwidth]{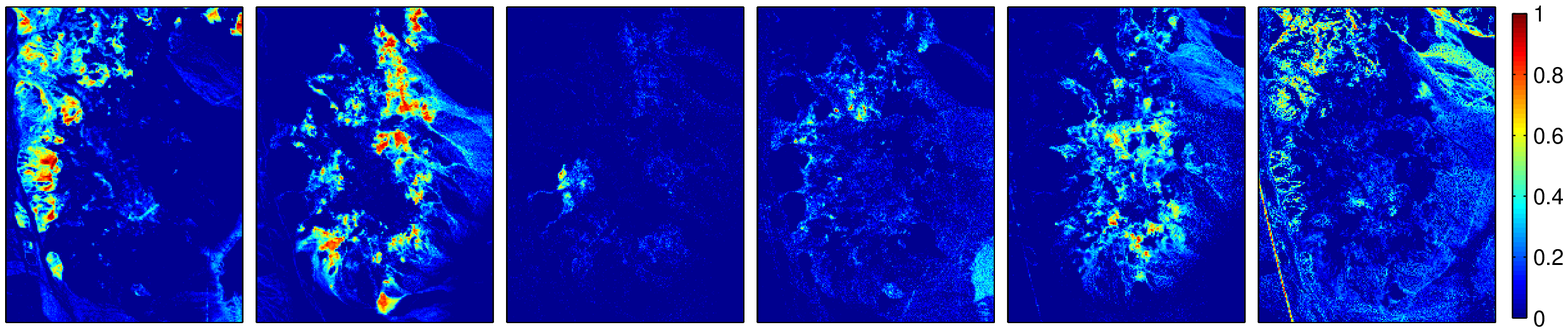}
\includegraphics[trim = 39mm 13mm 9mm 73mm, clip,width=0.85\textwidth]{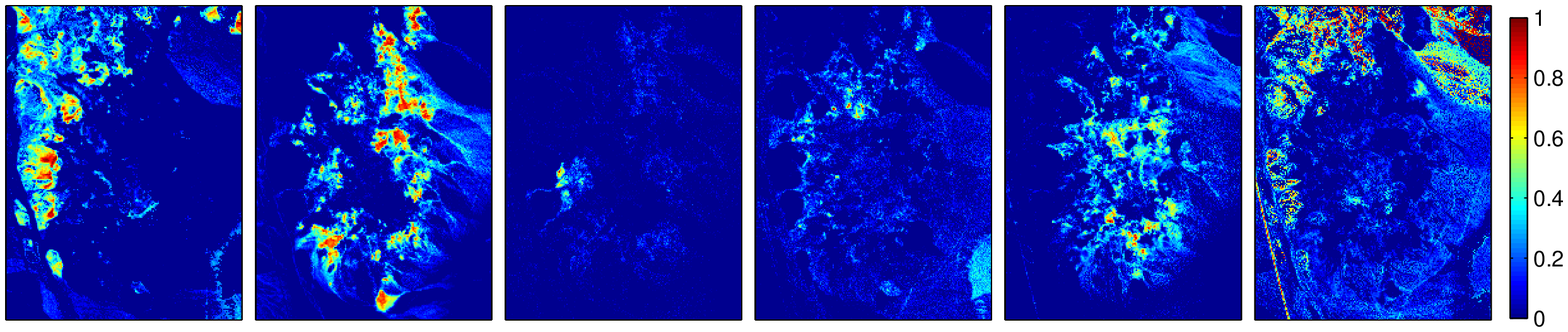}
\includegraphics[trim = 39mm 13mm 9mm 73mm, clip,width=0.85\textwidth]{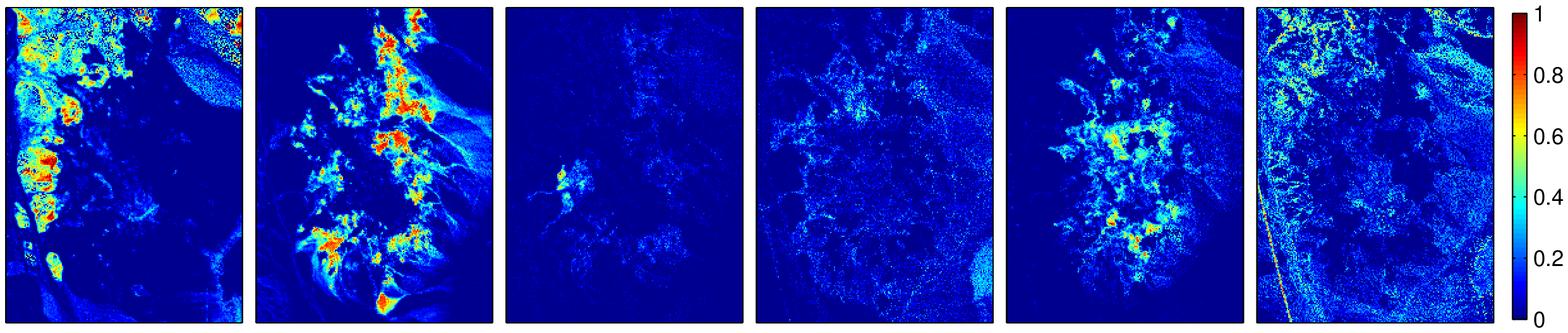}
\includegraphics[trim = 39mm 13mm 9mm 73mm, clip,width=0.85\textwidth]{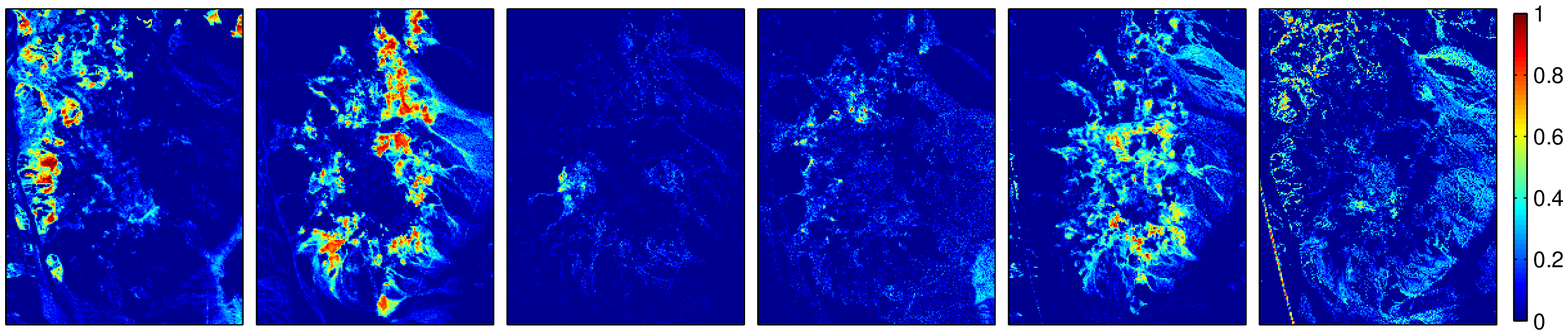}
\includegraphics[trim = 39mm 13mm 9mm 73mm, clip,width=0.85\textwidth]{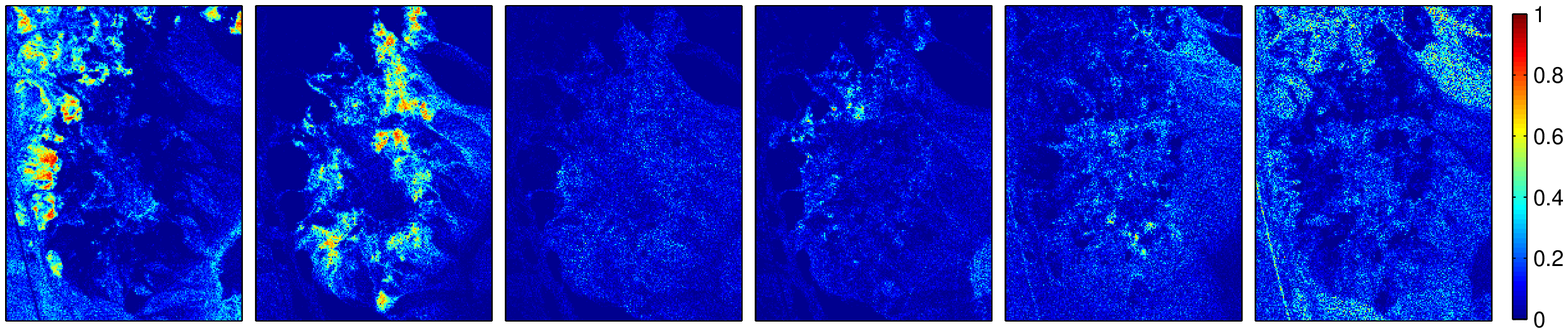}
\includegraphics[trim = 39mm 13mm 9mm 73mm, clip,width=0.85\textwidth]{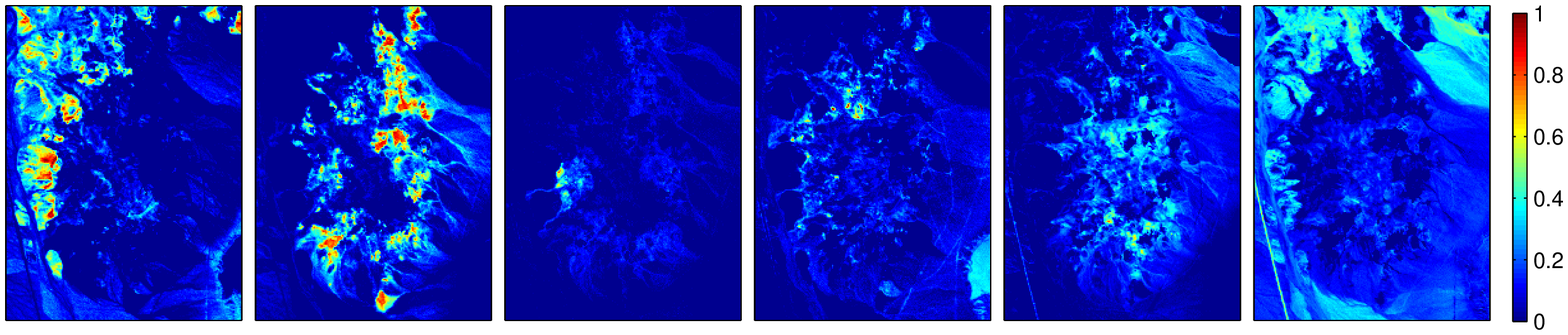}
\caption{\textbf{Cuprite image: } Estimated abundance maps using 187 clean bands. \textbf{Left to right:} sphene, alunite, buddingtonite, kaolinite, chalcedony, %US
highway% 95
. \textbf{Top to bottom:} SUnSAL-FCLS, FCLS, BayGBM, BayPPNMM, KFCLS, rNMF, CUSAL-FC.}
\label{Fig. AbunCuprite187}
\end{figure}

\begin{figure}
\centering
\graphicspath{{Graphics/Cuprite/}}
\centering
\small
\includegraphics[trim = 39mm 13mm 9mm 73mm, clip,width=0.85\textwidth]{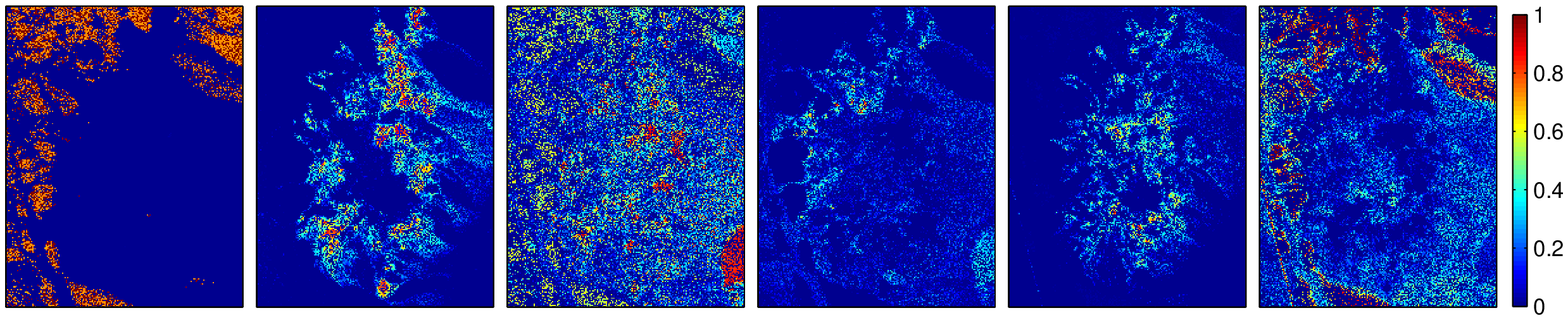}
\includegraphics[trim = 39mm 13mm 9mm 73mm, clip,width=0.85\textwidth]{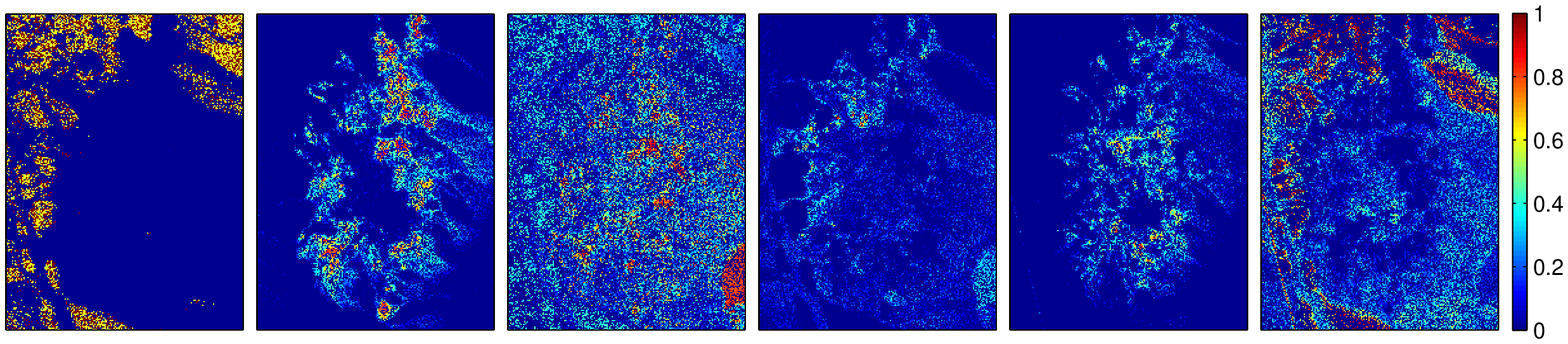}
\includegraphics[trim = 39mm 13mm 9mm 73mm, clip,width=0.85\textwidth]{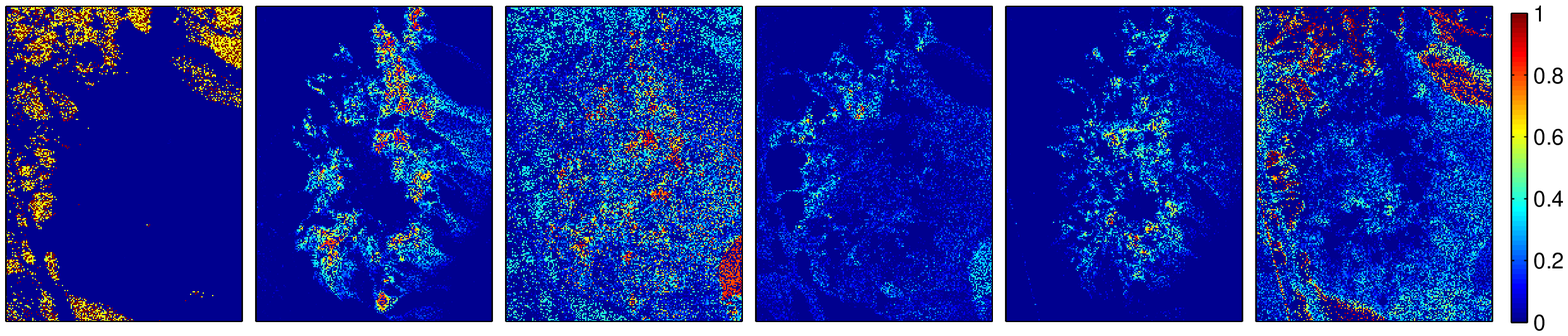}
\includegraphics[trim = 39mm 13mm 9mm 73mm, clip,width=0.85\textwidth]{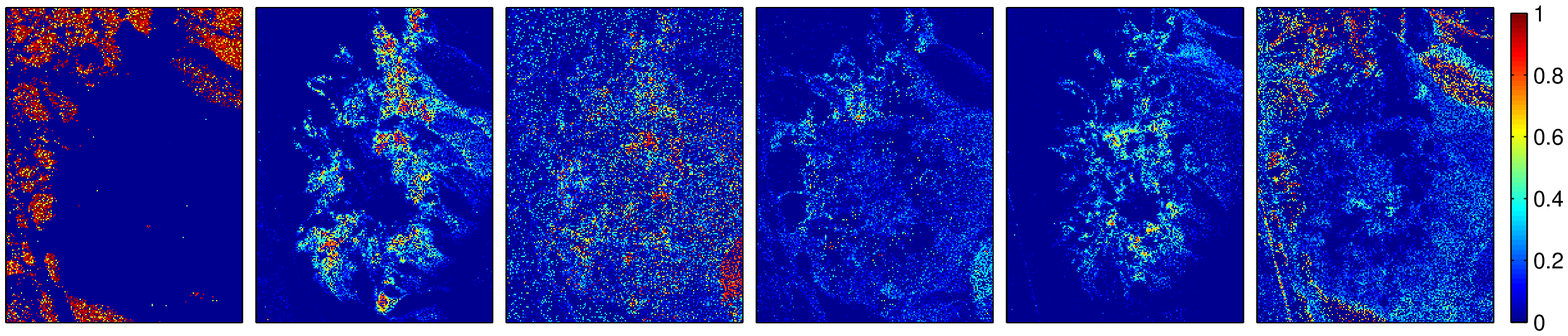}
\includegraphics[trim = 39mm 13mm 9mm 73mm, clip,width=0.85\textwidth]{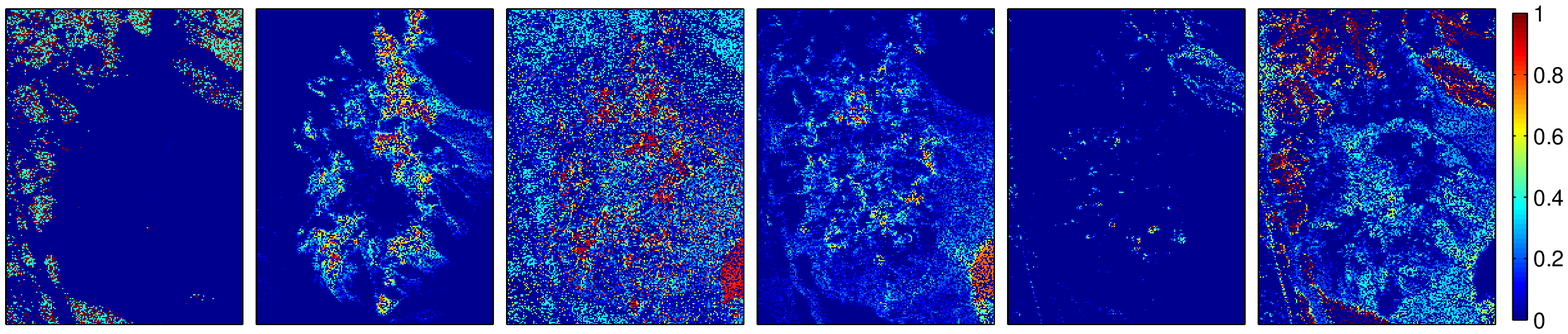}
\includegraphics[trim = 39mm 13mm 9mm 73mm, clip,width=0.85\textwidth]{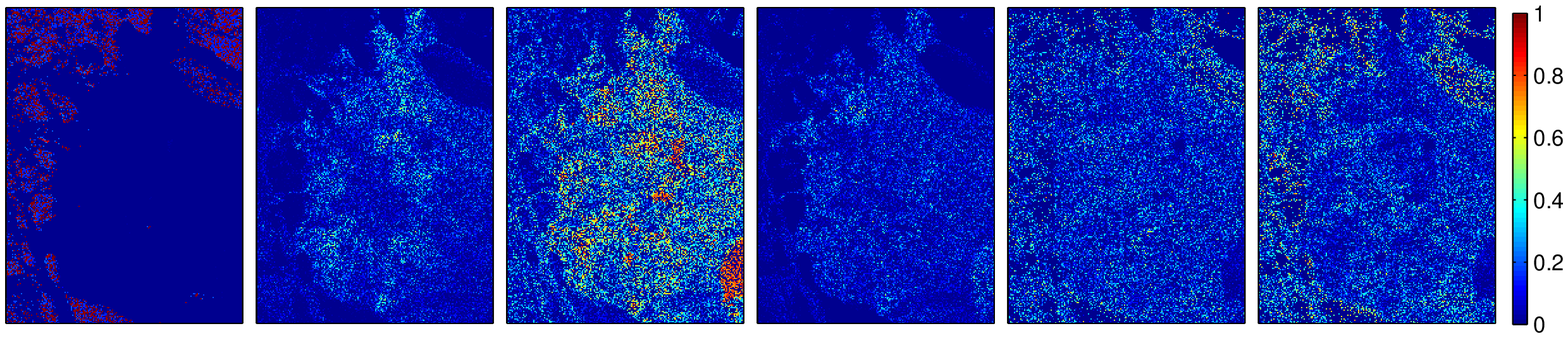}
\includegraphics[trim = 39mm 13mm 9mm 73mm, clip,width=0.85\textwidth]{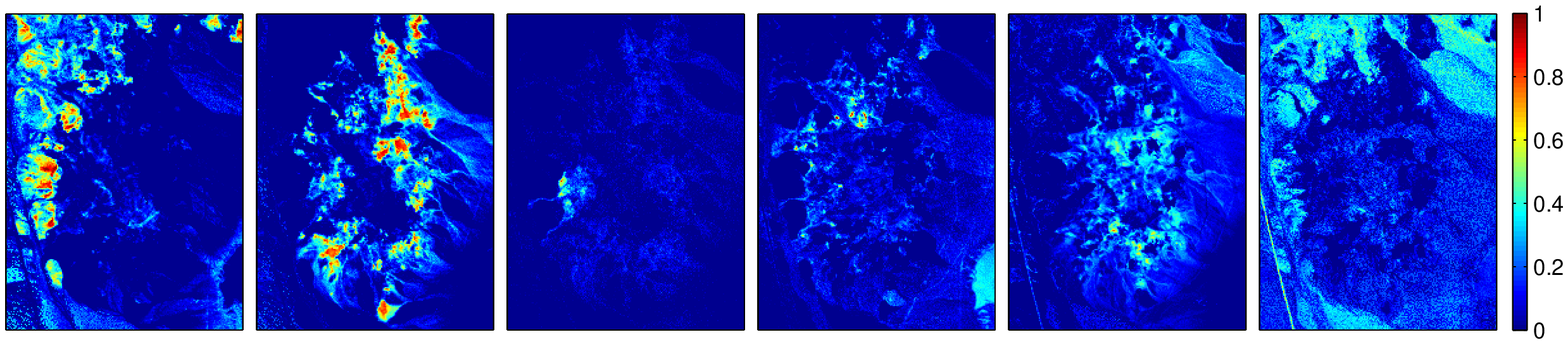}
\caption{\textbf{Cuprite image: } Estimated abundance maps using 205 bands, with 187 clean bands. Same legend as \figurename~\ref{Fig. AbunCuprite187}.}
\label{Fig. AbunCuprite205}
\end{figure}

\begin{figure}
\centering
\graphicspath{{Graphics/Cuprite/}}
\centering
\small
\includegraphics[trim = 39mm 13mm 9mm 73mm, clip,width=0.85\textwidth]{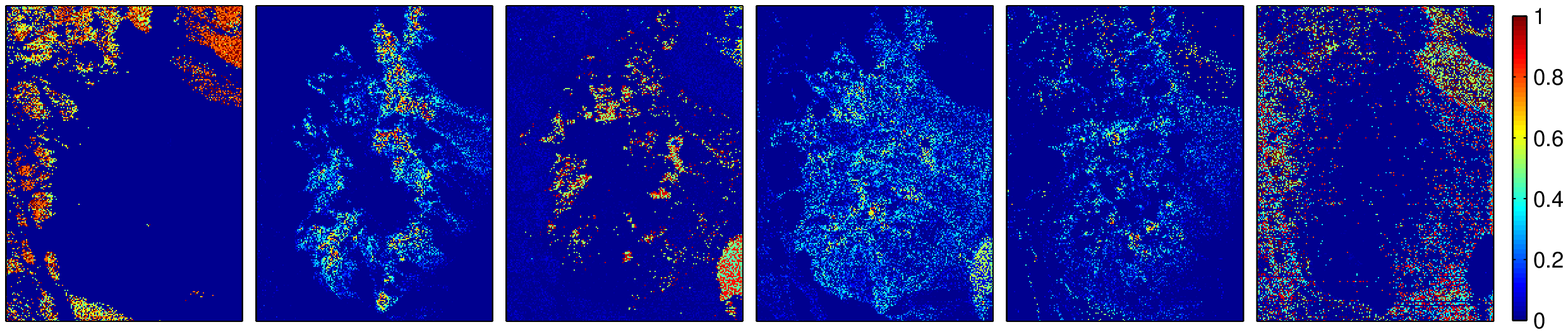}
\includegraphics[trim = 39mm 13mm 9mm 73mm, clip,width=0.85\textwidth]{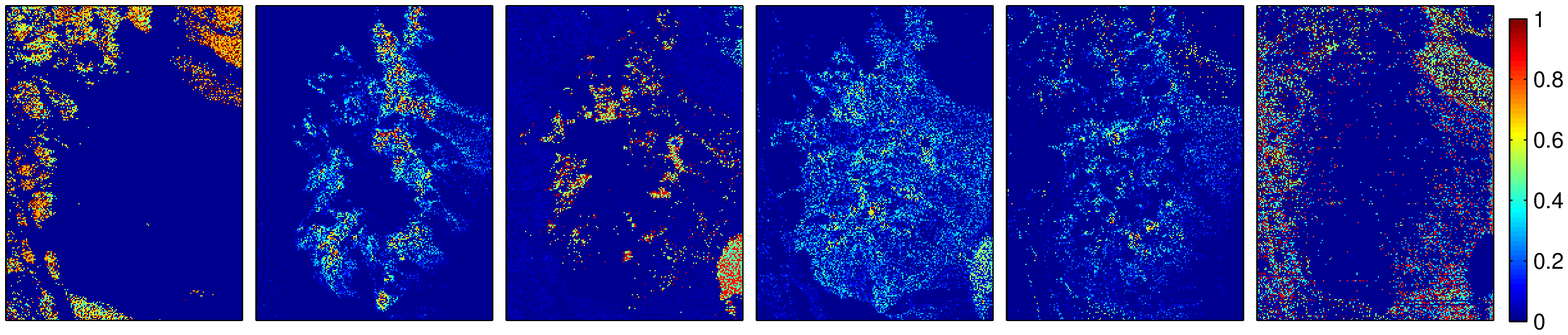}
\includegraphics[trim = 39mm 13mm 9mm 73mm, clip,width=0.85\textwidth]{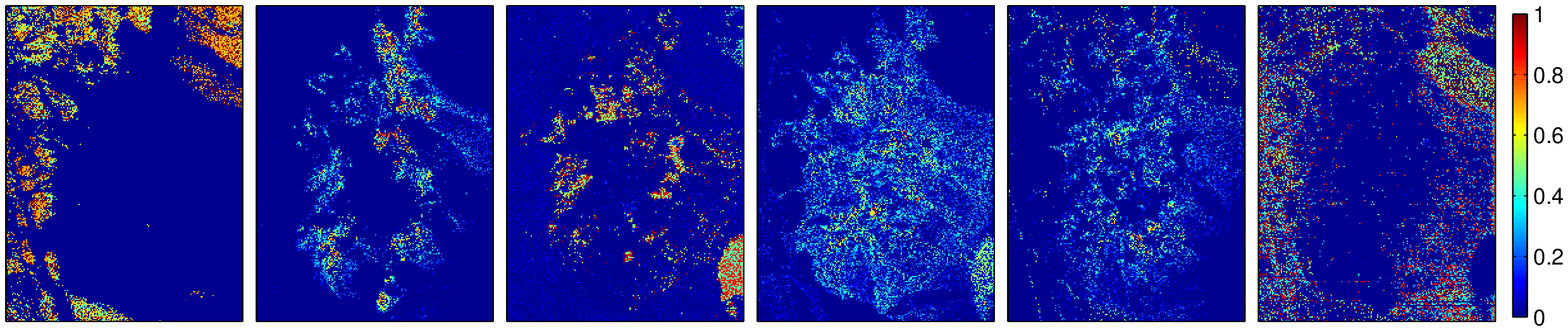}
\includegraphics[trim = 39mm 13mm 9mm 73mm, clip,width=0.85\textwidth]{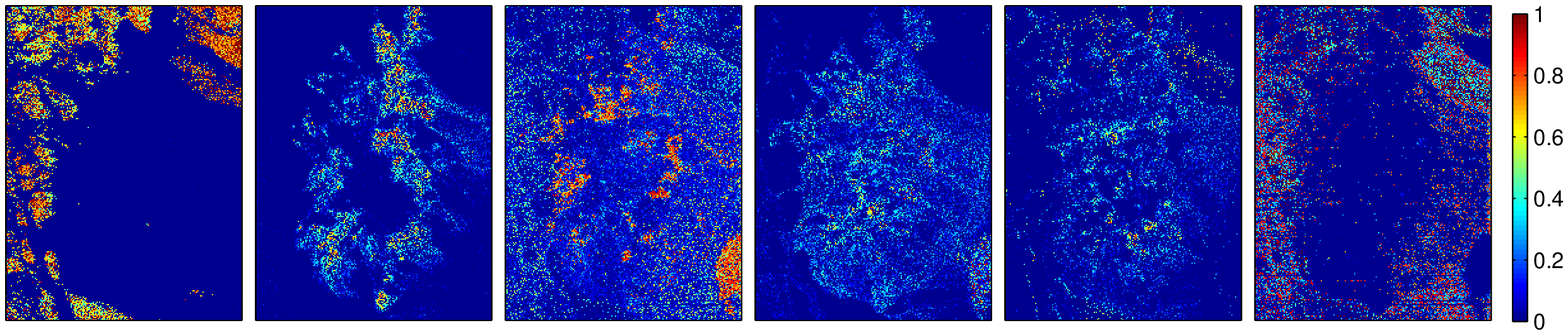}
\includegraphics[trim = 39mm 13mm 9mm 73mm, clip,width=0.85\textwidth]{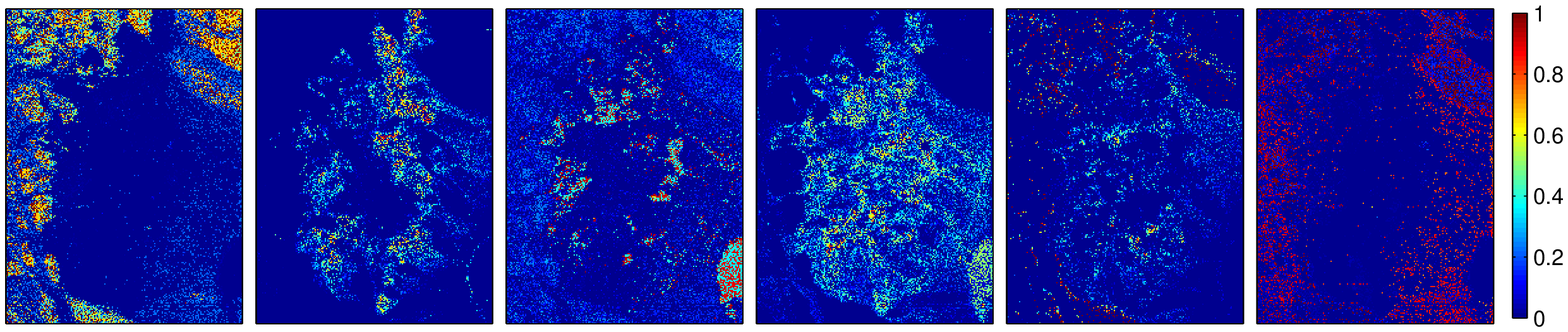}
\includegraphics[trim = 39mm 13mm 9mm 73mm, clip,width=0.85\textwidth]{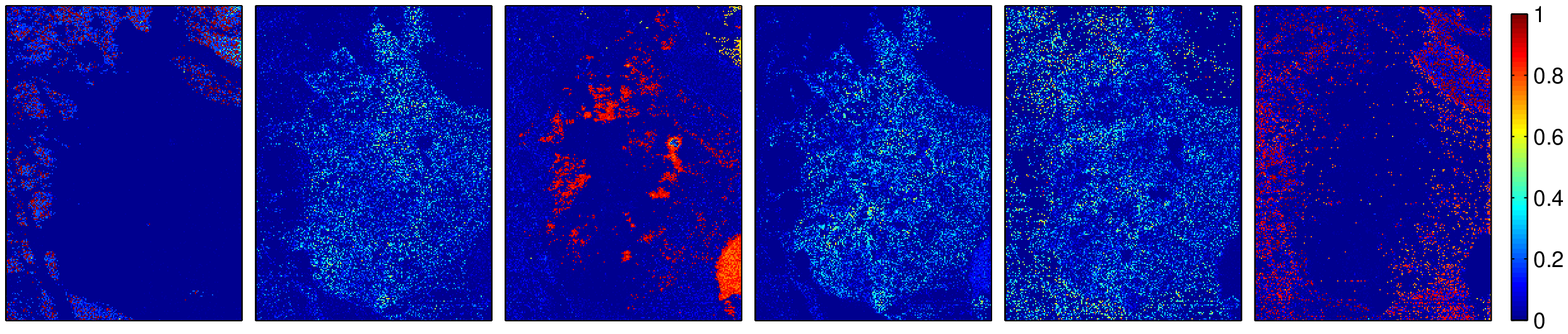}
\includegraphics[trim = 39mm 13mm 9mm 73mm, clip,width=0.85\textwidth]{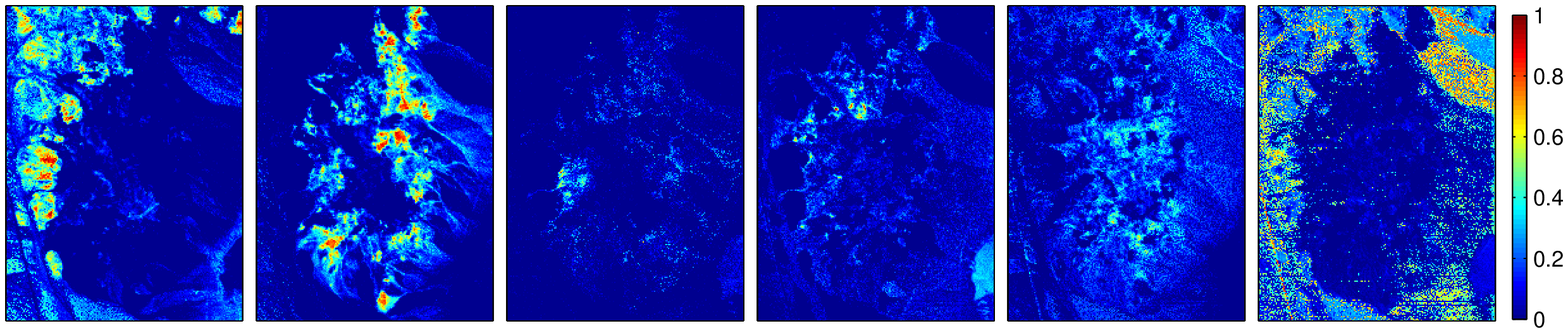}
\caption{\textbf{Cuprite image: } Estimated abundance maps using all the 224 bands, with 187 clean bands. Same legend as \figurename~\ref{Fig. AbunCuprite187}.}
\label{Fig. AbunCuprite224}
\end{figure}

Since ground-truth abundances are unknown, the performance is measured with the averaged spectral angle distance (SAD) between the input spectra ${\by}_{t}$ and the reconstructed ones $\widehat{\by}_{t}$, as illustrated in~\figurename~\ref{Fig.SADCuprite}%\footnote{The SAD is calculated without the noisy bands $1-3, 105-115, 150-170,$ and $223-224$.}
, where the SAD is defined by
\begin{equation*}
    \mathrm{SAD}=\frac{1}{T}\sum_{t=1}^T \arccos \Bigg( \frac{ \by_t^\top\widehat{\by}_{t}}{\|\by_t\|\| \widehat{\by}_{t} \|}\Bigg).%,
\end{equation*}
%where $\widehat{\by}_{t}$ is calculated corresponding to the unmixing mechanism of each algorithm.
%
%For fairness, as most of the aforementioned unmixing methods investigate the full-constraints, we consider the fully-constrained algorithm CUSAL-FC .
The estimated abundance maps using $187$, $205$ and $224$ bands are given in \figurename~\ref{Fig. AbunCuprite187}, \figurename~\ref{Fig. AbunCuprite205}, and \figurename~\ref{Fig. AbunCuprite224}, respectively.
In absence of noisy bands ({\em i.e.,} $L=187$ bands), all the considered methods lead to satisfactory abundance maps, with BayPPNMM providing the smallest SAD. As the number of noisy bands increases, especially from $L=199$ to $L=224$, the unmixing performance of the state-of-the-art methods deteriorates drastically, while the proposed CUSAL yields stable SAD. The obtained results confirm the good behavior of the proposed CUSAL algorithms and their robustness in presence of corrupted spectral bands.
% is robust to the noisy bands and

%% 187 clean bands

%% 205 bands

%% 224 bands

\section{Conclusion}\label{sec: Conclusion}

%This paper presented a supervised unmixing model by maximizing the correntropy, which is shown robustness to outliers. Taking the advantage of ADMM, we developed the algorithms for two cases, namely the fully-constrained and the sparsity-promoting versions. The effectiveness of the proposed methods were validated on unmixing both the synthetic and real hyperspectral images.
%However, due to the nonconvexity of the correntropy, the ADMM is slowed down to some degree by the bandwidth tuning step. Future works include the speed-up approach and the study on kernels.
This paper presented a supervised unmixing algorithm based on the correntropy maximization principle. %This criterion was shown to be robust to outliers and corrupted bands. %{\red
Two correntropy-based unmixing problems were addressed, the first with the non-negativity and sum-to-one constraints, and the second with the non-negativity constraint and a sparsity-promoting term.
The %abundance were estimated using an
alternating direction method of multipliers (ADMM) was investigated in order to solve the correntropy-based unmixing problems. The effectiveness and robustness of the proposed unmixing method were validated on synthetic and real hyperspectral images. 
Future works include the generalization of the correntropy criterion to account for the multiple reflection phenomenon~\cite{HaAlDoTo2011, fan2009comparative}, as well as incorporating nonlinear models~\cite{16.variability}.

% if have a single appendix:
%\appendix[Proof of the Zonklar Equations]
% or
%\appendix  % for no appendix heading
% do not use \section anymore after \appendix, only \section*
% is possibly needed

% use appendices with more than one appendix
% then use \section to start each appendix
% you must declare a \section before using any
% \subsection or using \label (\appendices by itself
% starts a section numbered zero.)
%

%\appendices
%\section{Proof of the First Zonklar Equation}
%Appendix one text goes here.
%
%% you can choose not to have a title for an appendix
%% if you want by leaving the argument blank
%\section{}
%Appendix two text goes here.

% use section* for acknowledgement
\section*{Acknowledgment}

This work was supported by the French ANR, grant \mbox{HYPANEMA: ANR-12BS03-0033}.\\

\bigskip\bigskip

% Can use something like this to put references on a page
% by themselves when using endfloat and the captionsoff option.
\ifCLASSOPTIONcaptionsoff
  \newpage
\fi

% trigger a \newpage just before the given reference
% number - used to balance the columns on the last page
% adjust value as needed - may need to be readjusted if
% the document is modified later
%\IEEEtriggeratref{8}
% The "triggered" command can be changed if desired:
%\IEEEtriggercmd{\enlargethispage{-5in}}

% references section

% can use a bibliography generated by BibTeX as a .bbl file
% BibTeX documentation can be easily obtained at:
% http://www.ctan.org/tex-archive/biblio/bibtex/contrib/doc/
% The IEEEtran BibTeX style support page is at:
% http://www.michaelshell.org/tex/ieeetran/bibtex/
%\bibliographystyle{IEEEtran}
% argument is your BibTeX string definitions and bibliography database(s)
%\bibliography{IEEEabrv,../bib/paper}
%
% <OR> manually copy in the resultant .bbl file
% set second argument of \begin to the number of references
% (used to reserve space for the reference number labels box)
%\begin{thebibliography}{1}
%
%\bibitem{IEEEhowto:kopka}
%H.~Kopka and P.~W. Daly, \emph{A Guide to \LaTeX}, 3rd~ed.\hskip 1em plus
%  0.5em minus 0.4em\relax Harlow, England: Addison-Wesley, 1999.
%
%\end{thebibliography}

\bibliographystyle{IEEEtran}
\bibliography{bib_fei,hyperspec} %biblio_ph
% biography section
%
% If you have an EPS/PDF photo (graphicx package needed) extra braces are
% needed around the contents of the optional argument to biography to prevent
% the LaTeX parser from getting confused when it sees the complicated
% \includegraphics command within an optional argument. (You could create
% your own custom macro containing the \includegraphics command to make things
% simpler here.)
%\begin{IEEEbiography}[{\includegraphics[width=1in,height=1.25in,clip,keepaspectratio]{mshell}}]{Michael Shell}
% or if you just want to reserve a space for a photo:

% You can push biographies down or up by placing
% a \vfill before or after them. The appropriate
% use of \vfill depends on what kind of text is
% on the last page and whether or not the columns
% are being equalized.

%\vfill

% Can be used to pull up biographies so that the bottom of the last one
% is flush with the other column.
%\enlargethispage{-5in}

% that's all folks
\end{document}